\documentclass[10pt,twocolumn,letterpaper]{article}

\usepackage[pagenumbers]{cvpr} 
\definecolor{cvprblue}{rgb}{0.21,0.49,0.74}
\usepackage{float}
\usepackage[pagebackref,breaklinks,colorlinks,allcolors=cvprblue]{hyperref}
\usepackage[all]{hypcap}

\usepackage{bm,multirow}
\usepackage{amsmath}
\usepackage{amssymb}
\usepackage{adjustbox}
\usepackage{pifont}

\def\paperID{4253} 
\def\confName{CVPR}
\def\confYear{2025}

\def\modelName{SwiftEdit}

\newcommand{\cmark}{\textcolor{green}{\ding{51}}} 
\newcommand{\xmark}{\textcolor{red}{\ding{55}}}   

\newcommand{\minisection}[1]{\vspace{2mm}\noindent{\textbf{#1}}}

\newcommand\blfootnote[1]{%
  \begingroup
  \renewcommand\thefootnote{}\footnote{#1}%
  \addtocounter{footnote}{-1}%
  \endgroup
}

\title{\modelName: Lightning Fast Text-Guided Image Editing via One-Step Diffusion}

\author{
Trong-Tung Nguyen, Quang Nguyen, Khoi Nguyen, Anh Tran, Cuong Pham$^\dagger$ \\
Qualcomm AI Research$^{+}$\\
{\tt\small \{tunnguy,quanghon,khoinguy,anhtra,pcuong\}@qti.qualcomm.com}}

\def\mD{\mathcal{D}}
\def\mE{\mathcal{E}}

\def\mN{\mathcal{N}}

\def\1n{\mathbf{1}_n}
\def\0{\mathbf{0}}
\def\1{\mathbf{1}}

\def\F{{\bf F}}
\def\G{{\bf G}}

\def\Z{{\bf Z}}

\def\c{{\bf c}}

\def\h{{\bf h}}

\def\x{{\bf x}}

\def\z{{\bf z}}

\def\bepsilon{\mbox{\boldmath{$\epsilon$}}}

\newcommand{\cm}[1]{}

\newcommand{\myheading}[1]{\vspace{1ex}\noindent \textbf{#1}}

\newif\ifshowsolution
\showsolutiontrue
\ifshowsolution

\else

\fi

\begin{document}


\makeatletter
\g@addto@macro\@maketitle{\vspace{-13mm}
  \begin{figure}[H]
  \setlength{\linewidth}{\textwidth}
  \setlength{\hsize}{\textwidth}
  \centering
  \includegraphics[width=.9\textwidth]{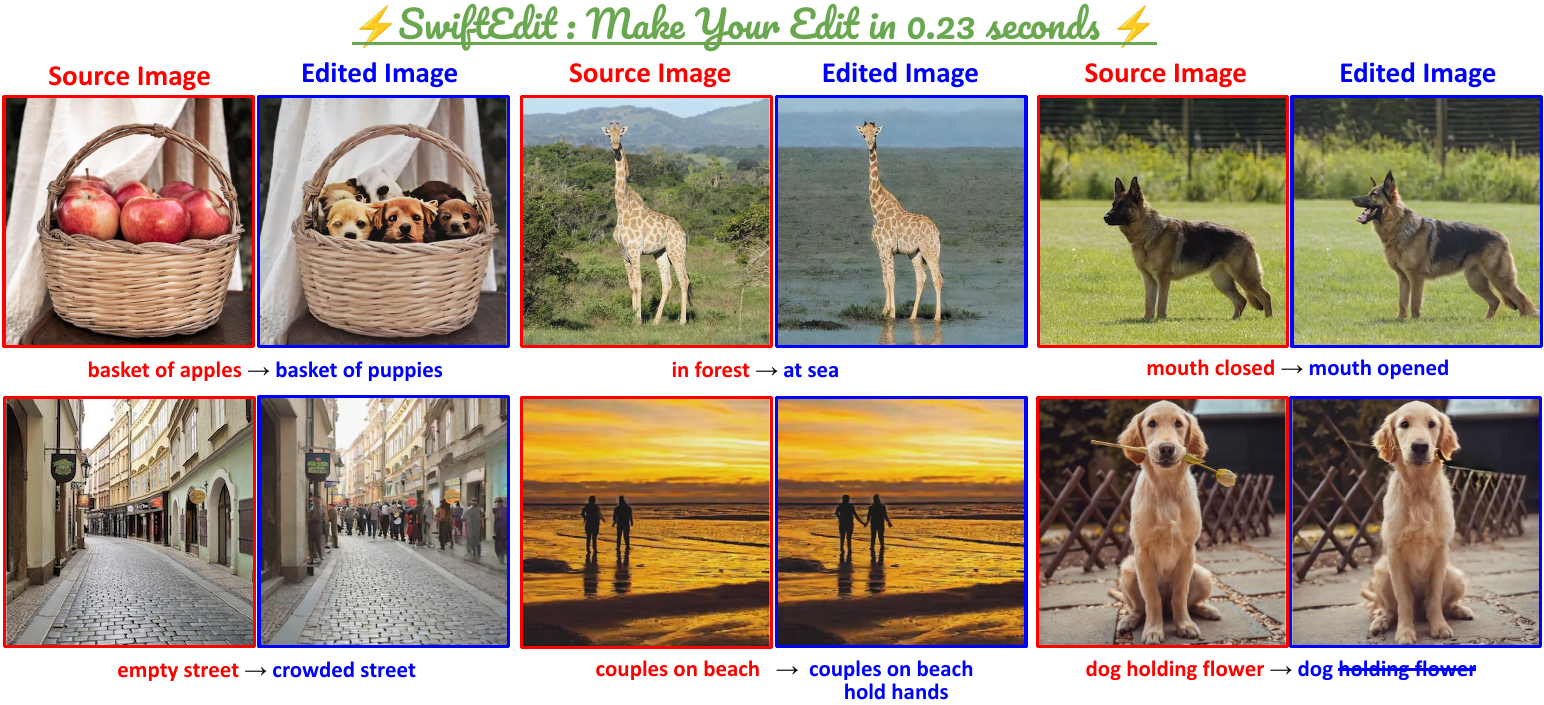}
  \vspace{-3mm}
  \caption{SwiftEdit empowers instant, localized image editing using only text prompts, freeing users from the need to define masks. In just 0.23 seconds on a single A100 GPU, it unlocks a world of creative possibilities demonstrated across diverse editing scenarios.} \label{fig:teaser}
  \end{figure}
}
\makeatother

\maketitle

\blfootnote{$^+$Qualcomm Vietnam Company Limited}
\blfootnote{$^\dagger$ also affiliated with Posts \& Telecom. Inst. of Tech., Vietnam}
\blfootnote{Contact email: \href{mailto:nguyentrongtung11101999@gmail.com}{nguyentrongtung11101999@gmail.com}}

\begin{abstract}
Recent advances in text-guided image editing enable users to perform image edits through simple text inputs, leveraging the extensive priors of multi-step diffusion-based text-to-image models. However, these methods often fall short of the speed demands required for real-world and on-device applications due to the costly multi-step inversion and sampling process involved. In response to this, we introduce SwiftEdit, a simple yet highly efficient editing tool that achieve instant text-guided image editing (\textbf{in 0.23s}). The advancement of SwiftEdit lies in its two novel contributions: a one-step inversion framework that enables one-step image reconstruction via inversion and a mask-guided editing technique with our proposed attention rescaling mechanism to perform localized image editing. Extensive experiments are provided to demonstrate the effectiveness and efficiency of SwiftEdit. In particular, SwiftEdit enables instant text-guided image editing, which is extremely faster than previous multi-step methods (at least \textbf{50$\times$ times faster}) while maintain a competitive performance in editing results. Our project is at \url{https://swift-edit.github.io/}.
\end{abstract}    
\vspace{-15pt}
\section{Introduction}
\label{sec:intro}
Recent text-to-image diffusion models \cite{NEURIPS2022_ec795aea, Rombach_2022_CVPR, podell2024sdxl, dao2025swiftbrush} have achieved remarkable results in generating high-quality images semantically aligned with given text prompts. To generate realistic images, most of them rely on multi-step sampling techniques, which reverse the diffusion process starting from random noise to realistic image. To overcome this time-consuming sampling process, some works focus on reducing the number of sampling steps to a few (4-8 steps) \cite{10.1145/3680528.3687625} or even one step \cite{yin2024onestep, yin2024improved, nguyen2024swiftbrush, dao2025swiftbrush} via distillation techniques while not compromising results. These approaches not only accelerate image generation but also enable faster inference for downstream tasks, such as image editing.

For text-guided image editing, recent approaches \cite{Mokady_2023_CVPR, ju2023direct, li2023stylediffusion} use an inversion process to determine the initial noise for a source image, allowing for (1) source image reconstruction and (2) content modification aligned with guided text while preserving other details. Starting from this inverted noise, additional techniques, such as attention manipulation and hijacking \cite{cao_2023_masactrl, Tumanyan_2023_CVPR, nguyen2024flexedit}, are applied at each denoising step to inject edits gradually while preserving key background elements. This typical approach, however, is resource-intensive, requiring two lengthy \textbf{multi-step} processes: inversion and editing. To address this, recent works \cite{10.1145/3680528.3687612, 10.1007/978-3-031-72630-9_23, starodubcev2024invertible} use \textbf{few-step} diffusion models, like SD-Turbo \cite{sauer2025adversarial}, to reduce the sampling steps required for inversion and editing, incorporating additional guidance for disentangled editing via text prompts. However, these methods still struggle to achieve sufficiently fast text-guided image editing for on-device applications while maintaining performance competitive with multistep approaches.

\begin{figure}[t]
    \centering
    \includegraphics[width=\columnwidth]{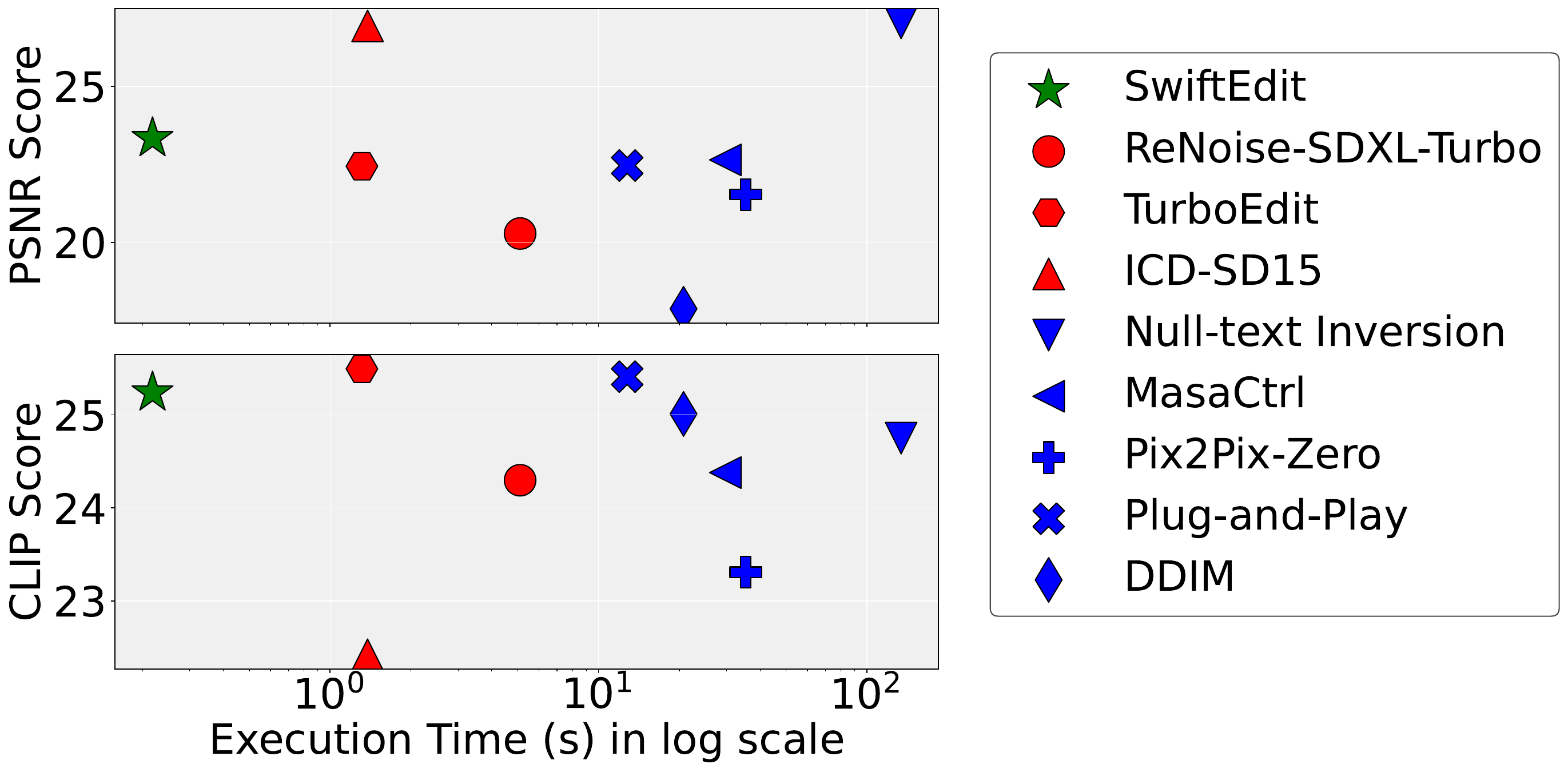}
    \vspace{-10pt}
    \caption{ 
        Comparing our \textcolor{OliveGreen}{one-step} SwiftEdit with \textcolor{red}{few-step} and \textcolor{blue}{multi-step diffusion} editing methods in terms of background preservation (PSNR), editing semantics (CLIP score), and runtime. Our method delivers lightning-fast text-guided editing while achieving competitive results.
    }
    \label{fig:plot_compare}
    \vspace{-12pt}
\end{figure}

In this work, we take a different approach by building on a \textbf{one-step} text-to-image model for image editing. We introduce SwiftEdit -- the first one-step text-guided image editing tool -- which achieves at least 50$\times$ faster execution than previous multi-step methods while maintaining competitive editing quality. Notably, both the inversion and editing in SwiftEdit are accomplished in a single step.

Inverting one-step diffusion models is challenging, as existing techniques like DDIM Inversion \cite{song2020denoising} and Null-text Inversion \cite{Mokady_2023_CVPR} are unsuitable for our one-step real-time editing goal. To achieve this, we design a novel one-step inversion framework inspired by encoder-based GAN Inversion methods \cite{zhu2020indomain, wang2021HFGI, 9792208}. Unlike GAN inversion, which requires domain-specific networks and retraining, our inversion framework generalizes to any input images. For this, we leverage SwiftBrushv2 \cite{dao2025swiftbrush}, a recent one-step text-to-image model known for speed, diversity, and quality, using it as both the \textbf{one-step image generator} and backbone for our \textbf{one-step inversion network}. We then train it with weights initialized from SwiftBrushv2 to handle any source inputs through a two-stage training strategy, combining supervision from both synthetic and real data.

Following the one-step inversion, we introduce an efficient mask-based editing technique. Our method can either accept an input editing mask or infer it directly from the trained inversion network and guidance prompts. The mask is then used in our novel attention-rescaling technique to blend and control the edit strength while preserving background elements, enabling high-quality editing results.

To the best of our knowledge, our work is the first to explore diffusion-based one-step inversion using a one-step text-to-image generation model to instantly perform text-guided image editing (\textbf{in 0.23 seconds}). While being significantly fast compared to other multi-step and few-step editing methods, our approach achieves a competitive editing result as shown in \cref{fig:plot_compare}. In summary, our main contribution includes:

\begin{itemize}
    \item We propose a novel one-step inversion framework trained with a two-stage strategy. Once trained, our framework can invert any input images into an editable latent in a single step without further retraining or finetuning.
    \item We show that our well-trained inversion framework can produce an editing mask guided by source and target text prompts within a single batchified forward pass.
    \item We propose a novel attention-rescaling technique for mask-based editing, offering flexible control over editing strength while preserving key background information.
\end{itemize}
\section{Related Work}
\label{sec:related_works}
\subsection{Text-to-image Diffusion Models}
Diffusion-based text-to-image models \cite{podell2024sdxl, NEURIPS2022_ec795aea, Rombach_2022_CVPR} typically rely on computationally expensive iterative denoising to generate realistic images from Gaussian noise. Recent advances \cite{salimans2022progressive, meng2023distillation, pmlr-v202-song23a, luo2023latent} alleviate this by distilling the knowledge from multi-step teacher models into a few-step student network. Notable works \cite{liu2023instaflow, pmlr-v202-song23a, luo2023latent, yin2024onestep, yin2024improved, nguyen2024swiftbrush, dao2025swiftbrush} show that this knowledge can be distilled even into a one-step student model. Specifically, Instaflow \cite{liu2023instaflow} uses rectified flow to train a one-step network, while DMD \cite{yin2024onestep} applies distribution-matching objectives for knowledge transfer. DMDv2 \cite{yin2024improved} removes costly regression losses, enabling efficient few-step sampling. SwiftBrush \cite{nguyen2024swiftbrush} utilizes an image-free distillation method with text-to-3D generation objectives, and SwiftBrushv2 \cite{dao2025swiftbrush} integrates post-training model merging and clamped CLIP loss, surpassing its teacher model to achieve state-of-the-art one-step text-to-image performance. These one-step models provide rich prior information about text-image alignment and are extremely fast, making them ideal for our one-step text-based image editing approach.

\subsection{Text-based Image Editing}
Several approaches leverage the strong prior of image-text relationships in text-to-image models to perform text-guided \textbf{multi-step} image editing via an inverse-to-edit approach. First, they invert the source image into ``informative" noise. Methods like DDIM Inversion \cite{song2020denoising} use linear approximations of noise prediction, while Null-text Inversion \cite{Mokady_2023_CVPR} enhances reconstruction quality through costly per-step optimization. Direct Inversion \cite{ju2023direct} bypasses these issues by disentangling source and target generation branches. Second, editing methods such as \cite{cao_2023_masactrl, Tumanyan_2023_CVPR, nguyen2024flexedit, 10.1145/3588432.3591513, hertz2023prompttoprompt} manipulate attention maps to embed edits while preserving background content. However, their multi-step diffusion process remains too slow for practical applications.

To address this issue, several works \cite{starodubcev2024invertible, 10.1007/978-3-031-72630-9_23, 10.1145/3680528.3687612} enable few-step image editing using fast generation models \cite{10.1145/3680528.3687625}. ICD \cite{starodubcev2024invertible} achieves accurate inversion in 3-4 steps with a consistency distillation framework, followed by text-guided editing. ReNoise \cite{10.1007/978-3-031-72630-9_23} refines the sampling process with an iterative renoising technique at each step. TurboEdit \cite{10.1145/3680528.3687612} uses a shifted noise schedule to align inverted noise with the expected schedule in fast models like SDXL Turbo \cite{10.1145/3680528.3687625}. Though these methods reduce inference time, they fall short of instant text-based image editing needed for fast applications. Our one-step inversion and one-step localized editing approach dramatically boosts time efficiency while surpassing few-step methods in editing performance.

\subsection{GAN Inversion}
GAN inversion \cite{zhu2020indomain, Perarnau2016, wang2021HFGI, lipton2017precise, creswell2018inverting, ma2018invertibility, bau2019seeing}
maps a source image into the latent space of a pre-trained GAN, allowing the generator to recreate the image, which is valuable for tasks like image editing. Effective editing requires a latent space that can both reconstruct the image and support realistic edits through variations in the latent code. Approaches fall into three groups: encoder-based \cite{Perarnau2016, zhu2020indomain, zhu2016generative}, optimization-based \cite{lipton2017precise, creswell2018inverting, ma2018invertibility}, and hybrid \cite{bau2019seeing, bau2019inverting, zhu2020indomain}. Encoder-based methods learn a mapping from the image to the latent code for fast reconstruction. Optimization-based methods refine a code by iteratively optimizing it, while hybrid methods combine both, using an encoder’s output as initialization for further optimization. Inspired by encoder-based speed, we develop a one-step inversion network, but instead of GAN, we leverage a one-step text-to-image diffusion model. This allows us to achieve text-based image editing across diverse domains rather than being restricted to specific domain as in GAN-based methods.
\section{Preliminaries}
\label{sec:preliminaries}
\begin{figure*}[ht]
    \centering
    \includegraphics[width=\textwidth]{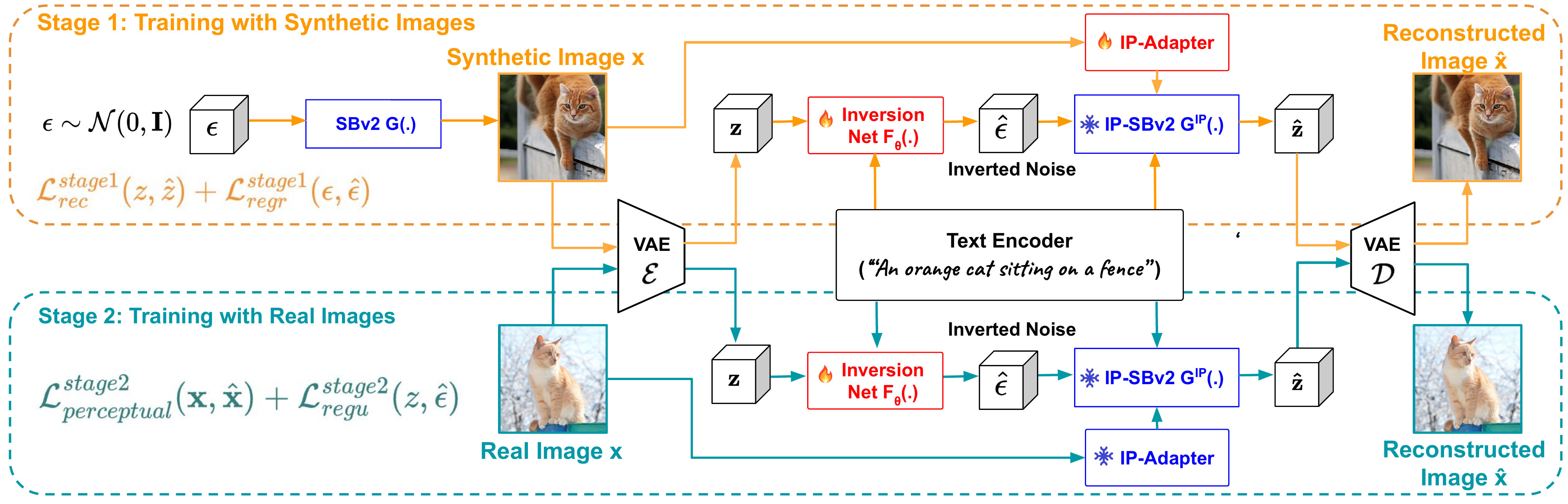}
    \caption{\textbf{Proposed two-stage training for our one-step inversion framework}. In stage 1, we warms up our inversion network on synthetic data generated by SwiftBrushv2. At stage 2, we shift our focus to real images, continue to train our inversion network to enable instantly image inversion for any input images without additional fine-tuning or retraining.}
    \label{fig:diagram}
    \vspace{-5pt}
\end{figure*}
\vspace{-5pt}
\myheading{Multi-step diffusion model.}
Text-to-image diffusion model $\bepsilon_\phi$ attempts to generate image $\hat{\x}$ given the target prompt embedding $\c_y$ (extracted from the CLIP text encoder of a given text prompt $y$)  through a $T$ iterative denoising steps, starting from Gaussian noise, $\z_T = \bepsilon \sim \mathcal{N}(0,I)$:
\begin{equation}
\z_{t-1} = \frac{\z_t - \sigma_t \bepsilon_\phi(\z_t, t, \c_y)}{\alpha_t} + \delta_t \bepsilon_t, \quad \bepsilon_t \sim \mathcal{N}(0,I),
\end{equation}
where $t$ is the timestep, and $\sigma_t, \alpha_t, \delta_t$ are three coefficients. 
The final latent $\z = \z_0$  is then input to a VAE decoder $\mD$ to produce the image $\hat{\x}=\mD(\z)$. 

\minisection{One-step diffusion model}.
The traditional diffusion model’s sampling process requires multiple steps, making it time-consuming. To address this, one-step text-to-image diffusion models like InstaFlow \cite{liu2023instaflow}, DMD \cite{yin2024onestep}, DMD2 \cite{yin2024improved}, SwiftBrush \cite{nguyen2024swiftbrush}, and SwiftBrushv2 \cite{dao2025swiftbrush} have been developed, reducing the sampling steps to a single step. Specifically, one-step text-to-image diffusion model $\G$ aims to transform a noise input $\bepsilon \sim \mathcal{N}(0,1)$, given a text prompt embedding $\c_y$, directly into an image latent $\hat{\z}$,  without iterative denoising steps, or $\hat{\z} = \G(\bepsilon, \c_y)$. SwiftBrushv2 (SBv2) stands out in one-step image generation by quickly producing high-quality, diverse outputs, forming the basis of our approach. Building on its predecessor, SBv2 integrates key improvements: it uses SD-Turbo initialization for enhanced output quality, a clamped CLIP loss to strengthen visual-text alignment, and model fusion with post-enhancement techniques, all contributing to superior performance and visual fidelity.

\minisection{Score Distillation Sampling (SDS)} \cite{poole2023dreamfusion} is a popular objective function that utilizes the strong prior learned by 2D diffusion models to optimize a target data point $\z$ by calculating its gradient as follows:
\begin{equation}
\nabla_{\theta} \mathcal{L}_{\text{SDS}}\triangleq \mathbb{E}_{t, \bepsilon} \left[w(t)\left(\bepsilon_\phi(\z_t, t, \c_y)  - \bepsilon\right) \frac{\partial \z}{ \partial \theta}\right],
\label{eq:sdsgrad}
\end{equation}
where $\z=g(\theta)$ is rendered by a differentiable image generator $g$ parameterized by $\theta$, $\z_t$ denotes a perturbed version of $\z$ with a random amount of noise $\bepsilon$, and $w(t)$ is a scaling function corresponding to the timestep $t$.
The objective of SDS loss is to provide an updated direction that would move $\z$ to a high-density region of the data manifold using the score function of the diffusion model $\bepsilon_\phi(\z_t, t, \c_y)$. Notably, this gradient omits the Jacobian term of the diffusion backbone, removing the expensive computation when backpropagating through the entire diffusion model U-Net.

\minisection{Image-Prompt via Decoupled Cross-Attention}. 
IP-Adapter \cite{ye2023ip-adapter} introduces an image-prompt condition $\x$ that can be seamlessly integrated into a pre-trained text-to-image generation model. It achieves this through a decoupled cross-attention mechanism, which separates the conditioning effects of text and image features. This is done by adding an extra cross-attention layer to each cross-attention layer in the original U-Net. Given image features $\c_\x$ (extracted from $\x$ by a CLIP image encoder), text features $\c_y$ (from text prompt $y$ using a CLIP text encoder), and query features $\Z_l$ from the previous U-Net layer $l-1$, the output $\h_l$ of the decoupled cross-attention is computed as:
\begin{align}
\h_{l} &= \operatorname{Attn}(Q_l, K_y, V_y) + s_{\x}\operatorname{Attn}(Q_l, K_\x, V_\x)
\label{eq:decoupled_cross_map},
\end{align}
where $\operatorname{Attn}(.)$ denotes the attention operation. Scaling factors $s_{\x}$ is used to control the influence of $\c_\x$ on the generated output. $Q_l = W^Q \Z_l$ is the query matrix projected by the weight matrix $W^Q$. The key and value matrices for text features $\c_y$ are $K_y = W^K_y \c_y$ and $V_y = W^V_y \c_y$, respectively, while the projected key and value matrices for image features $\c_\x$ are $K_\x = W^K_\x \c_\x$ and $V_\x = W^V_\x \c_\x$. Notably, only the two weight matrices $W^K_\x$ and $W^V_\x$ are trainable, while the remaining weights remain frozen to preserve the original behavior of the pretrained diffusion model.

\section{Proposed Method}
\label{sec:proposed_method}
Our goal is to enable instant image editing with the one-step text-to-image model, SBv2. In \cref{sec:two_stage_training}, we develop a one-step inversion network that predicts inverted noise to reconstruct a source image when passed through SBv2. We introduce a \textbf{two-stage training strategy} for this inversion network, enabling single-step reconstruction of any input images without further retraining. An overview is shown in \cref{fig:diagram}. During inference, as described in \cref{sec:editing_scale}, we use self-guided editing mask to locate edited regions. Our attention-rescaling technique then utilizes the mask to achieve disentangled editing and control the editing strength while preserving the background.

\subsection{Inversion Network and Two-stage Training}
\label{sec:two_stage_training}
Given an input image that may be synthetic (generated by a model like SBv2) or real, our first objective is to inverse and reconstruct it using SBv2 model. To achieve this, we develop a one-step inversion network $\F_\theta$ to transform the image latent $\z$ into an inverted noise $\hat{\bepsilon}=\F_\theta(\z, \c_y)$, and then feed back to SBv2 to compute the reconstructed latent $ \hat{\z} = \G(\hat{\bepsilon}, \c_y) = \G(\F_\theta(\z, \c_y), \c_y).$
For synthetic images, training $\F_\theta$ is straightforward, with pairs $(\bepsilon, \z)$, where $\bepsilon$ is the noise used to generate $\z$, allowing direct regression of $\hat{\bepsilon}$ to $\bepsilon$, and aligning the inverted noise with SBv2’s input noise distribution. However, for real images, the domain gap poses a challenge, as the original noise $\bepsilon$ is unavailable, preventing us from computing regression objective and potentially causing $\hat{\bepsilon}$ to deviate from the desired distribution. In the following section, we discuss our inversion network and a two-stage training strategy designed to overcome these challenges effectively.

\myheading{Our Inversion Network $\F_\theta$} follows the architecture of the one-step diffusion model $\G$ and is initialized with $\G$’s weights. However, we found this approach suboptimal: the inverted noise $\hat{\bepsilon}$ predicted by $\F_\theta$ attempts to perfectly reconstruct the input image, leading to overfitting on specific patterns from the input. This tailoring makes the noise overly dependent on input features, which limits editing flexibility.

To overcome this, we introduce an auxiliary, image-conditioned branch -- similar to IP-Adapter \cite{ye2023ip-adapter} -- within the one-step generator $\G$, named $\G^{\text{IP}}$. This branch integrates image features encoded from the input image $\x$ along with text prompt $y$, aiding in reconstruction and reducing the need for $\F_\theta$ to embed extensive visual details from the input image. This approach effectively alleviates the burden on $\hat{\bepsilon}$, enhancing both reconstruction and editing capabilities. We compute the inverted noise $\hat{\bepsilon}$ along with the reconstructed image latent $\hat{\z}$ as follows:
\begin{equation}
    \hat{\bepsilon} = \F_\theta(\z, c_y), \quad \hat{\z} = \G^{\text{IP}}(\hat{\bepsilon}, \c_y, \c_{\x}).
\end{equation}

\begin{figure}[ht]
    \centering
    \includegraphics[trim={0.5cm 1cm 2.5cm 1cm}, width=\columnwidth]{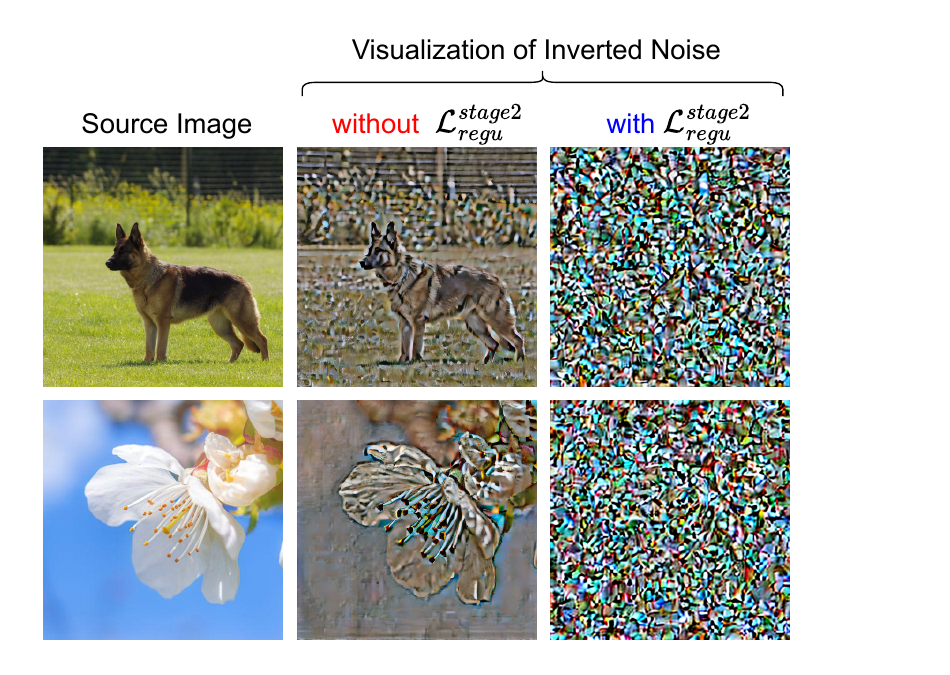}
    \caption{Comparison of inverted noise predicted by our inversion network when trained without and with stage 2 regularization loss.}
    \label{fig:compare_noise}
    \vspace{-10pt}
\end{figure}
\myheading{Stage 1: Training with synthetic images.}
As mentioned above, this stage aims to pretrain the inversion network $\F_\theta$ with synthetic training data sampled from a text-to-image diffusion network  $\G$, i.e., SBv2. In \cref{fig:diagram}, we visualize the flow of stage 1 training in \textcolor{orange}{orange color}. Pairs of training samples $(\bepsilon,\z)$ are created as follows:
\begin{equation}
    \bepsilon \sim \mathcal{N}(0, 1), \quad \z = \G(\bepsilon, \c_y).
\end{equation}
We combine the reconstruction loss $\mathcal{L}^{\text{stage1}}_{\text{rec}}$ and regression loss $\mathcal{L}^{\text{stage1}}_{\text{regr}}$ to train the inversion network $\F_\theta$ and part of the IP-Adapter branch (including the linear mapping and cross-attention layers for image conditions). The regression loss $\mathcal{L}^{\text{stage1}}_{\text{regr}}$ encourages $\F_\theta(.)$ to produce an inverted noise $\hat{\bepsilon}$ that closely follows SBv2's input noise distribution by regressing $\hat{\bepsilon}$ to $\bepsilon$. This ensures that the inverted noise remains close to the multivariate normal distribution, which is crucial for effective editability as shown in prior work \cite{Mokady_2023_CVPR}. On the other hand, the reconstruction loss $\mathcal{L}^{\text{stage1}}_{\text{rec}}$ enforces alignment between the reconstructed latent $\hat{\z}$ and the original source latent $\z$, preserving input image details. In summary, the training objectives are as follows:
\begin{align}
    \mathcal{L}^{\text{stage1}}_{\text{rec}} = ||\z - \hat{\z}||^2_2, \quad \mathcal{L}^{\text{stage1}}_{\text{regr}} = ||\bepsilon - \hat{\bepsilon}||^2_2,
\end{align}
\begin{equation}
\mathcal{L}^{\text{stage1}} = \mathcal{L}_{\text{rec}}^{\text{stage1}} + \lambda^{\text{stage1}}.\mathcal{L}_{\text{regr}}^{\text{stage1}},
\label{eq:loss_stage1}
\end{equation}
where we set $\lambda^{\text{stage1}}=1$ during training. After this stage, our inversion framework could reconstruct source input images generated by the SBv2 model. However, it fails to work with real images due to the domain gap which motivates us to continue training with stage 2. 

\minisection{Stage 2: Training with real images.}
We replace the reconstruction loss from stage 1 with a perceptual loss using the Deep Image Structure and Texture Similarity (DISTS) metric \cite{9298952}. This perceptual loss, $\mathcal{L}^{\text{stage2}}_{\text{perceptual}} = \operatorname{DISTS}(\x, \hat{\x})$, compares $\hat{\x} = \mD(\hat{\z})$ (where $\hat{\z}=\G^{\text{IP}}(\hat{\bepsilon}, \c_y, \c_{\x})$) with the real input image $\x$. DISTS is trained on real images, capturing perceptual details in structure and texture, making it a more robust visual similarity measure than the pixel-wise reconstruction loss used in stage 1.

Since the original noise $\bepsilon$, used to reconstruct $\z$ in SBv2, is unavailable at this stage, we cannot directly apply the regression objective from stage 1. Training stage 2 solely with $\mathcal{L}^{\text{stage2}}_{\text{perceptual}}$ can cause the inverted noise $\hat{\bepsilon}$ to drift from the ideal noise distribution $\mN(0, I)$, as the perceptual loss encourages $\hat{\bepsilon}$ to capture source image patterns, aiding reconstruction but constraining future editing flexibility (see \cref{fig:compare_noise}, column 2). To address this, we introduce a new regularization term $\mathcal{L}_{\text{regu}}^{\text{stage2}}$, inspired by Score Distillation Sampling (SDS) as defined in \cref{eq:sdsgrad}. The SDS gradient steers the optimized latent toward dense regions of the data manifold. Given that the real image latent $\z = \mE(\x)$ already lies in a high-density region, we shift the optimization focus to the noise term $\bepsilon$, treating our inverted noise as an added noise to $\z$. We then compute the loss gradient as follows:
\begin{gather}
    \hat{\bepsilon} = \F_\theta(\z, \c_y), \quad \z_t = \alpha_t \z + \sigma_t \hat{\bepsilon}, \nonumber \\
    \nabla_{\theta} \mathcal{L}_{\text{regu}}^{\text{stage2}}\triangleq \mathbb{E}_{t, \hat{\bepsilon}} \left[w(t)\left(\hat{\bepsilon}-\bepsilon_\phi(\z_t, t, \c_y)\right)\frac{\partial \hat{\bepsilon}}{\partial \theta}\right].
\label{eq:regugrad}
\end{gather}
Our regularization gradient has the opposite sign of \cref{eq:sdsgrad} since it optimizes $\hat{\bepsilon}$ instead of $\z$ (derivation details in Appendix). After initializing from stage 1, $\hat{\boldsymbol{\epsilon}}$ resembles Gaussian noise $ \mN(0, 1) $, making the noisy latent $ \z_t $ compatible with the multi-step teacher's training data. This allows the teacher to accurately predict $\bepsilon_\phi(\z_t, t, \c_y)$, and achieve $ \boldsymbol{\epsilon}_\phi(\z_t, t, \c_y) - \hat{\boldsymbol{\epsilon}} \approx \mathbf{0} $. Thus, $\hat{\bepsilon}$ stays the same. Over time, the reconstruction loss nudges $\F_\theta$ to generate an inverted noise, $\hat{\boldsymbol{\epsilon}}$, tailored for reconstruction, diverging from $ \mathcal{N}(0, 1) $ and creating an unfamiliar $ \mathbf{z}_t $. The resulting gradient prevents excessive drift from the original distribution, reinforcing stability from stage 1, as shown in third column of \cref{fig:compare_noise}.
Similar to stage 1, we combine both perceptual losses $\mathcal{L}^{\text{stage2}}_{\text{perceptual}}$ and regularization loss $\mathcal{L}^{\text{stage2}}_{\text{regu}}$ where we set $\lambda^\text{stage2}=1$. During training , we train only the inversion network, keeping the IP-Adapter branch and decoupled cross-attention layers frozen to retain the image prior features learned in stage 1. Flow of training stage 2 are visualized as \textcolor{teal}{\bf teal color} in \cref{fig:diagram}.
\begin{figure}[ht]
    \centering
    \begin{subfigure}[b]{\columnwidth}
        \centering
        \includegraphics[trim={1cm 1cm 2cm 0}, width=1.\columnwidth]{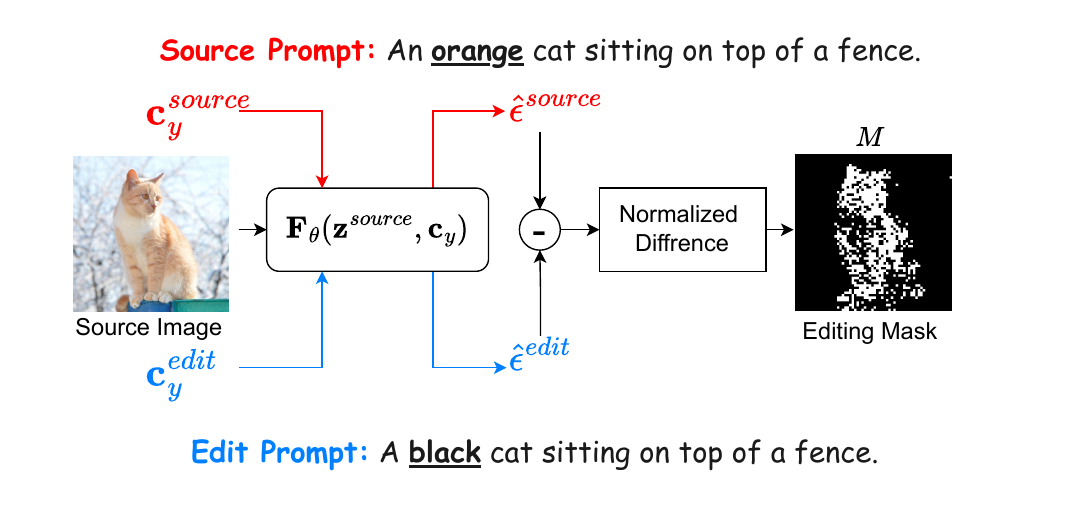}
        \caption{\textbf{Self-guided editing mask extraction}. Given source and editing prompts, our inversion network predicts two different noise maps, highlighting the editing regions $M$.}
        \label{fig:mask_extract}
    \end{subfigure}
    \begin{subfigure}[b]{\columnwidth}
        \centering
        \includegraphics[width=\columnwidth]{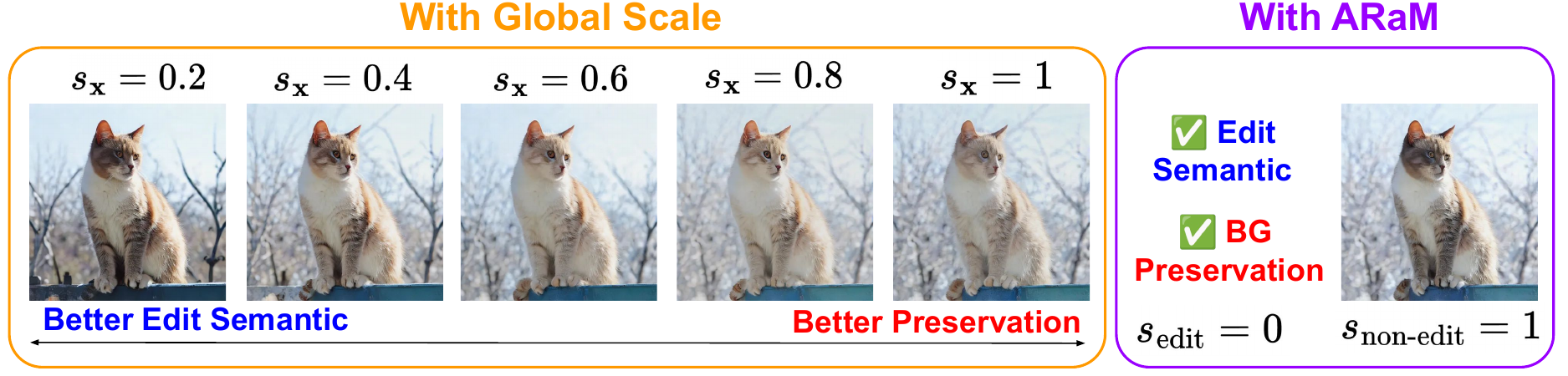}
        \caption{\textbf{Effect of global scale and our edit-aware scale}. Comparison of edited results between varying global image condition scale $s_{\x}$ with our ARaM.}
        \label{fig:vary_scale}
    \end{subfigure} 
    \begin{subfigure}[b]{\columnwidth}
        \centering
        \includegraphics[width=\columnwidth]{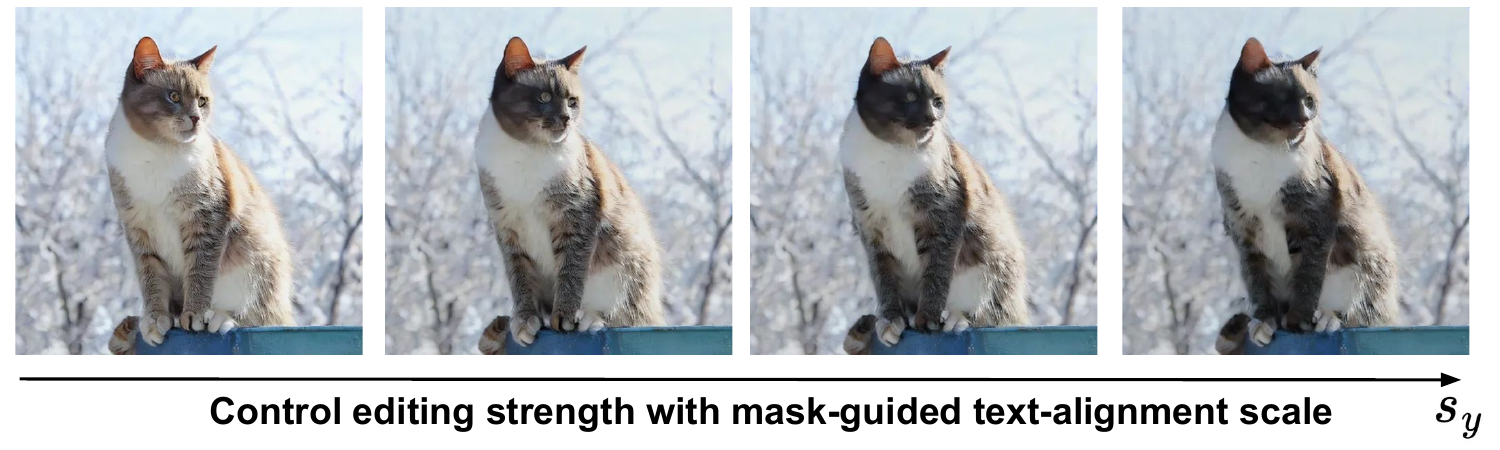}
        \caption{\textbf{Effect of editing strength scale}. Visualization of edited results when varying mask-based text-alignment scale $s_{y}$.}
        \label{fig:vary_edit_str}
    \end{subfigure}
    \vspace{-5mm}
    \caption{\textbf{Illustration of Attention Rescaling for Mask-aware Editing (ARaM)}. We apply attention rescaling with our self-guided editing mask to achieve local image editing and enable editing strength control.}
    \vspace{-4mm}
    \label{fig:edit_fig}
\end{figure}
\subsection{Attention Rescaling for Mask-aware Editing (ARaM)}
\label{sec:editing_scale}
During inference, given a source image $\x^{\text{source}}$, a source prompt $y^{\text{source}}$, and an editing prompt $y^{\text{edit}}$, our target is to produce an edited image $\x^{\text{edit}}$ following the editing prompt without modifying irrelevant background elements. After two-stage training, we obtain a well-trained inversion network $\F_\theta$ to transform source image latent $\z^{\text{source}}=\mE(\x^{\text{source}})$ to inverted noise $\hat{\bepsilon}$. Intuitively, we can use the one-step image generator, $\G^{\text{IP}}(.)$, to regenerate the image but with an edit prompt embedding $\c_y^\text{edit}$ as guided prompt instead. The edited image latent is computed via $\z^{\text{edit}}=\G^{\text{IP}}(\hat{\bepsilon}, \c_y^\text{edit}, \c_{\x})$. 
As discussed in \cref{sec:two_stage_training}, the source image condition $\c_\x$ is crucial for reconstruction, with its influence modulated by $s_{\x}$ as shown in \cref{eq:decoupled_cross_map}. To illustrate this, we vary $s_{\x}$ while generating the edited image $\x^{\text{edit}}=\mD(\z^{\text{edit}})$ in \textcolor{orange}{orange block} of \cref{fig:vary_scale}. As shown, higher values of $s_{\x}$ enforce fidelity to the source image, limiting editing flexibility due to tight control by $\c_x$. Conversely, lower $s_{\x}$ allows more flexible edits but reduces reconstruction quality. Based on this observation, we introduce Attention Rescaling for Mask-aware editing (ARaM) in $\G^{\text{IP}}$, guided by the editing mask $M$. The key idea is to amplify the influence of $\c_{\x}$ in non-edited regions for better preservation while reducing its effect within edited regions, providing greater editing flexibility. To implement this, we reformulate the computation in \cref{eq:decoupled_cross_map} within $\G^{\text{IP}}$ by removing the global scale $s_{\x}$ and introducing region-specific scales as follows:
\begin{equation}
\begin{split}
    \h_{l} = &\:  s_{y}.M.\operatorname{Attn}(Q_l, K_y, V_y) \\ &
    +s_{\text{edit}}.M.\operatorname{Attn}(Q_l, K_\x, V_\x) \\ &
    +s_{\text{non-edit}}.(1-M).\operatorname{Attn}(Q_l, K_\x, V_\x).
\label{eq:ta_scale}
\end{split}
\end{equation}
This disentangled cross-attention differs slightly from \cref{eq:decoupled_cross_map} in  three scaling factors: $s_{y}$, $s_{\text{edit}}$, and $s_{\text{non-edit}}$, apply on different image regions. Two scaling factors $s_{\text{edit}}$, and $s_{\text{non-edit}}$ are used to separately control the influence of the image condition $\c_{\x}$ on the editing and non-editing regions. As shown in \textcolor{violet}{violet block} of \cref{fig:vary_scale}, this effectively results in an edited image which both follow prompt edit semantics and achieve good background preservation compared to using the same $s_\x$. On the other hand, we introduce the additional $s_{y}$ to lessen/strengthen the edit prompt-alignment effect within the editing region $M$ which could be used to control the editing strength as shown in \cref{fig:vary_edit_str}.
\begin{table*}[ht]
\centering
\small
\setlength{\tabcolsep}{8pt}
\begin{tabular}{llccccc} 
\toprule
\multirow{2}{*}{\textbf{Type}} & \multirow{2}{*}{\textbf{Method}} & \multicolumn{2}{c}{\textbf{Background Preservation}} & \multicolumn{2}{c}{\textbf{CLIP Semantics}} & \multirow{2}{*}{\textbf{Runtime}$\downarrow$} \\ 
\cmidrule(lr){3-4} \cmidrule(lr){5-6}
& & PSNR$\uparrow$ & MSE$_{\times 10 ^4}$$\downarrow$ & Whole $\uparrow$ & Edited$\uparrow$ & (seconds) \\ 
\midrule
\multirow{8.9}{*}{\begin{tabular}[c]{@{}l@{}}\textbf{Multi-step}\\\textbf{(50 steps)}\end{tabular}} & DDIM + P2P & 17.87 & 219.88 & 25.01 & 22.44 & 25.98 \\
& NT-Inv + P2P & 27.03 & 35.86 & 24.75 & 21.86 & 134.06 \\ 
\cmidrule{2-7}
& DDIM + MasaCtrl & 22.17 & 86.97 & 23.96 & 21.16 & 23.21 \\
& Direct Inversion + MasaCtrl & 22.64 & 81.09 & 24.38 & 21.35 & 29.68 \\ 
\cmidrule{2-7}
& DDIM + P2P-Zero & 20.44 & 144.12 & 22.80 & 20.54 & 35.57 \\
& Direct Inversion + P2P-Zero & 21.53 & 127.32 & 23.31 & 21.05 & 35.34 \\ 
\cmidrule{2-7}
& DDIM + PnP & 22.28 & 83.64 & 25.41 & 22.55 & 12.62 \\
& Direct Inversion + PnP & 22.46 & 80.45 & 25.41 & 22.62 & 12.79 \\ 
\midrule
\multirow{3}{*}{\begin{tabular}[c]{@{}l@{}}\textbf{Few-steps}\\\textbf{(4 steps)}\end{tabular}} & ReNoise (SDXL Turbo) & 20.28 & 54.08 & 24.29 & 21.07 & 5.11 \\
& TurboEdit & 22.43 & 9.48 & 25.49 & 21.82 & 1.32 \\
& ICD (SD 1.5) & 26.93 & 3.32 & 22.42 & 19.07 & 1.62 \\ 
\midrule
\multirow{2}{*}{\begin{tabular}[c]{@{}l@{}}\textbf{One-step}\\\end{tabular}}
& SwiftEdit (Ours) & 23.33 & 6.60 & 25.16 & 21.25 & \textbf{0.23} \\
& SwiftEdit (Ours with GT masks) & 23.31 & 6.18 & 25.56 & 21.91 & \textbf{0.23} \\
\bottomrule
\end{tabular}
\vspace{-5pt}
\caption{Quantitative comparison of SwiftEdit against other editing methods with metrics employed from PieBench \cite{ju2023direct}.}.
\label{tab:main_quant}
\vspace{-10pt}
\end{table*}

The editing mask $M$ discussed above can either be provided by the user or generated automatically from our inversion network $\F_\theta$. To extract \textbf{self-guided editing mask}, we observe that a well-trained $\F_\theta$ can discern spatial semantic differences in the inverted noise maps when conditioned on varying text prompts. As shown in \cref{fig:mask_extract}, we input the source image latent $\z^{\text{source}}$ to $\F_\theta$ with two different text prompts: the source $\c_y^{\text{source}}$ and the edit $\c_y^{\text{edit}}$. The difference noise map, $\hat{\bepsilon}^{\text{source}} - \hat{\bepsilon}^{\text{edit}}$, is then computed and normalized, yielding the editing mask $M$, which effectively highlights the editing areas.

\section{Experiments}
\subsection{Experimental Setup}

\vspace{-5pt}
\minisection{Dataset and evaluation metrics.}
We evaluate our editing performance on PieBench \cite{ju2023direct}, a popular benchmark containing 700 samples across 10 diverse editing types. Each sample includes a source prompt, edit prompt, instruction prompt, source image, and a manually annotated editing mask. Using PieBench's metrics, we assess both background preservation and editing semantics, aiming for a balance between them for high-quality edits. Background preservation is evaluated with PSNR and MSE scores on unedited regions of the source and edited images. Editing alignment is assessed using CLIP-Whole and CLIP-Edited scores, measuring prompt alignment with the full image and edited region, respectively.

\myheading{Implementation details.}
Our inversion network is based on the architecture of SBv2, initialized with SBv2 weights for stage 1 training. In stage 2, we continue training from stage 1's pretrained weights. For image encoding, we adopt the IP-Adapter \cite{ye2023ip-adapter} design, using a pretrained CLIP image encoder followed by a small projection network that maps the image embeddings to a sequence of features with length $N=4$, matching the text feature dimensions of the diffusion model. Both stages use the Adam optimizer \cite{DBLP:journals/corr/KingmaB14} with weight decay of 1e-4, a learning rate of 1e-5, and an exponential moving average (EMA) in every iteration. In stage 1, we train with a batch size of 4 for 100k iterations on synthetic samples generated by SBv2, paired with 40k captions from the JourneyDB dataset \cite{sun2024journeydb}. For stage 2, we train with a batch size of 1 and train over 180k iterations using 5k real images and their prompt descriptions from the CommonCanvas dataset \cite{gokaslan2023commoncanvas}. All experiments are conducted on a single NVIDIA A100 40GB GPU.

\begin{figure*}[ht]
    \centering
    \includegraphics[width=0.8\textwidth]{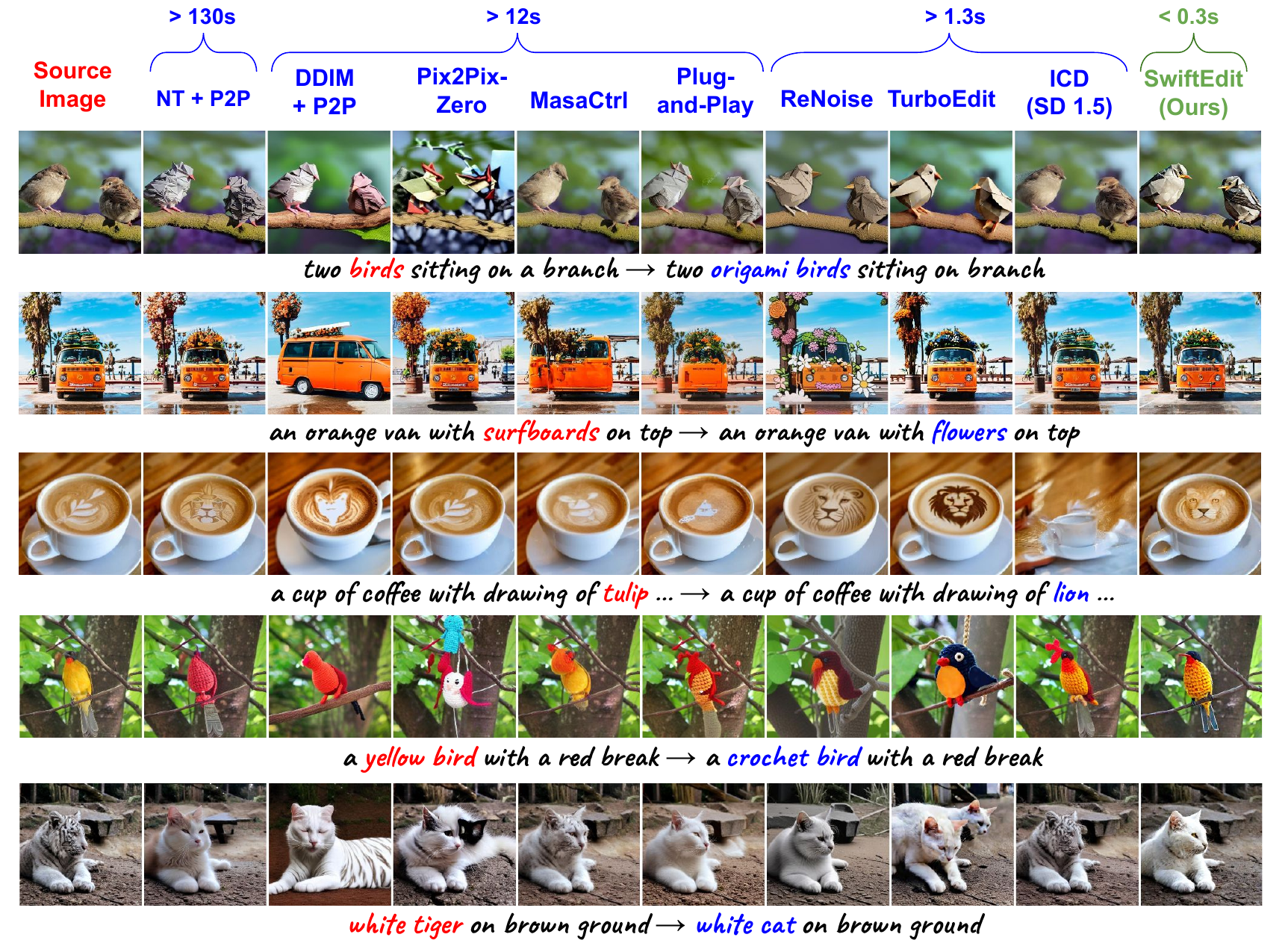}
    \vspace{-5pt}
    \caption{\textbf{Comparative edited results}. The first column shows the source image, while source and edit prompts are noted under each row.}
    \label{fig:main_qual}
    \vspace{-10pt}
\end{figure*}

\minisection{Comparison Methods.} We perform an extensive comparison of SwiftEdit with representative multi-step and recently introduced few-step image editing methods. For multi-step methods, we choose Prompt-to-Prompt (P2P) \cite{hertz2023prompttoprompt}, MasaCtrl \cite{cao_2023_masactrl}, Pix2Pix-Zero (P2P-Zero) \cite{10.1145/3588432.3591513}, and Plug-and-Play \cite{Tumanyan_2023_CVPR}, combined with corresponding inversion methods such as DDIM \cite{song2020denoising}, Null-text Inversion (NT-Inv) \cite{Mokady_2023_CVPR}, and Direct Inversion \cite{ju2023direct}. For few-step methods, we select Renoise \cite{10.1007/978-3-031-72630-9_23}, TurboEdit \cite{10.1145/3680528.3687612}, and ICD \cite{starodubcev2024invertible}.

\vspace{5pt}

\subsection{Comparison with Prior Methods}

\begin{figure}[t]
    \centering
    \includegraphics[width=\columnwidth]{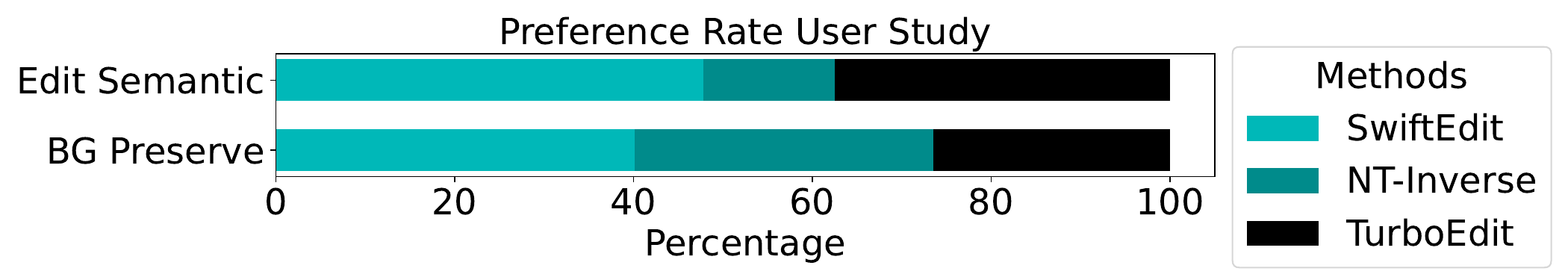}
    \caption{\textbf{User Study}.}
    \label{fig:user_study}
    \vspace{-12pt}
\end{figure}

\begin{table}[t]
\centering
\footnotesize
\setlength{\tabcolsep}{2pt}
\begin{tabular}{lcccc}
\toprule
\multirow{1}{*}{\textbf{Method}} 
& \textbf{PSNR}$\uparrow$ & \textbf{LPIPS}$_{\times 10 ^{3}}$$\downarrow$ & \textbf{MSE}$_{\times 10 ^{4}}$$\downarrow$ & \textbf{SSIM}$_{\times 10 ^{2}}$$\uparrow$ \\
\midrule
w/o stage 1 & 22.26 & 111.57 & 7.03 & 72.39  \\
 w/o stage 2 & 17.95 & 305.23 & 17.46 & 55.97 \\
w/o IP-Adapter  & 18.57 & 165.78 & 16.11 & 63.87\\
\midrule
Full Setting (Ours)&\textbf{24.35} & \textbf{89.69}& \textbf{4.59}&\textbf{76.34} 
  
\\
\bottomrule
\end{tabular}
\vspace{-5pt}
\caption{Impact of inversion framework design on real image reconstruction.}
\label{tab:ablation_framework_design}
\vspace{-10pt}
\end{table}

\begin{table}[t]
    \centering
    \small
    \setlength{\tabcolsep}{3pt}
    \begin{tabular}{lcccccc}
        \toprule
        \multirow{2}{*}{\textbf{Setting}} & \multirow{2}{*}{\textbf{$\mathcal{L}_{regr}^{stage1}$}} & \multirow{2}{*}{\textbf{$\mathcal{L}_{regu}^{stage2}$}} & \multicolumn{2}{c}{\textbf{CLIP Semantics}} \\ 
        \cmidrule(lr){4-5}
        & & & Whole ($\uparrow$) & Edited($\uparrow$) \\ 
        \midrule
        Setting 1 & \xmark & \xmark & 22.91 & 19.07 \\
        Setting 2 & \xmark & \cmark & 22.98 & 19.01 \\
        Setting 3 & \cmark & \xmark & 24.19 & 20.55 \\
        Setting 4 (Full) & \cmark & \cmark & \textbf{25.16} & \textbf{21.25} \\
        \bottomrule
    \end{tabular}
    \vspace{-5pt}
    \caption{Effect of loss on editing semantics score.}
    \label{tab:ablation_loss}
    \vspace{-15pt}
\end{table}
\vspace{-5pt}
\minisection{Quantitative Results.} In \cref{tab:main_quant}, we present the quantitative results comparing SwiftEdit to various multi-step and few-step image editing methods. Overall, SwiftEdit demonstrates superior time efficiency due to our one-step inversion and editing process, while maintaining competitive editing performance. Compared to multi-step methods, SwiftEdit shows strong results in background preservation scores, surpassing most approaches. Although it achieves a slightly lower PSNR score than NT-Inv + P2P, it has a better MSE score and is approximately 500 times faster. In terms of CLIP Semantics, we also achieve competitive results in CLIP-Whole (second best) and CLIP-Edited. Compared with few-step methods, SwiftEdit performs as the second-best in background preservation (with ICD being the best) and second-best in CLIP Semantics (with TurboEdit leading), while maintaining a speed advantage, being at least 5 times faster than these methods. Since SwiftEdit allows for user-defined editing masks, we also report results using the ground-truth editing masks from PieBench \cite{ju2023direct}. As shown in the last row of \cref{tab:main_quant}, results with the ground-truth masks show slight improvements, indicating that our self-guided editing masks are nearly as accurate as the ground truth.


\minisection{Qualitative Results.} In \cref{fig:main_qual}, we present visual comparisons of editing results generated by SwiftEdit and other methods. As illustrated, SwiftEdit successfully adheres to the given edit prompt while preserving essential background details. This balance demonstrates SwiftEdit’s strength over other multi-step methods, as it produces high-quality edits while being significantly faster.
When compared to few-step methods, SwiftEdit demonstrates a clear advantage in edit quality. Although ICD \cite{starodubcev2024invertible} scores high on background preservation (as shown in \cref{tab:main_quant}), it often fails to produce edits that align with the prompt. TurboEdit \cite{10.1145/3680528.3687612}, while achieving a higher CLIP score than SwiftEdit, generates lower-quality results that compromise key background elements, as seen in the first, second, and fifth rows of \cref{fig:main_qual}. This highlights SwiftEdit’s high-quality edits with prompt alignment and background preservation.


\minisection{User Study.} We conducted a user study with 140 participants to evaluate preferences for different editing results. Using 20 random edit prompts from PieBench \cite{ju2023direct}, participants compared images edited by three methods: Null-text Inversion \cite{Mokady_2023_CVPR}, TurboEdit \cite{10.1145/3680528.3687612}, and our SwiftEdit. Participants selected the most appropriate edits based on background preservation and editing semantics. As shown in \cref{fig:user_study}, SwiftEdit was the preferred choice, with 47.8\% favoring it for editing semantics and 40\% for background preservation, while also surpassing other methods in speed.

\section{Ablation Study}
\vspace{-10pt}
\myheading{Analysis of Inversion Framework Design.}
%
We conduct ablation studies to evaluate the impact of our inversion framework and two-stage training on image reconstruction. Our two-stage strategy is essential for the one-step inversion framework's effectiveness. In \cref{tab:ablation_framework_design}, we show that omitting any stages degrades reconstruction quality. The IP-Adapter with decoupled cross-attention is critical; removing it leads to poor reconstruction, as seen in row 3.

\myheading{Effect of loss on Editing Quality.}
As noted by \cite{Mokady_2023_CVPR}, an editable noise should follow a normal distribution to ensure flexibility. We conduct ablation studies to assess the impact of our loss functions on noise editability. As shown in \cref{tab:ablation_loss}, omitting any loss component reduces editability, measured by CLIP Semantics, while using both yields the highest scores. This emphasizes the importance of each loss in maintaining noise distributions that enhance editability.
\section{Conclusion and Discussion}
\label{sec:conclusion}
\vspace{-10pt}

\minisection{Conclusion.} In this work, we introduce SwiftEdit, a lightning-fast text-guided image editing tool capable of instant edits in 0.23 seconds. Extensive experiments demonstrate SwiftEdit’s ability to deliver high-quality results while significantly surpassing previous methods in speed, enabled by its one-step inversion and editing process. We hope SwiftEdit will facilitate interactive image editing.

\minisection{Discussion.} While \modelName~achieves instant-level image editing, challenges remain. Its performance still relies on the quality of the SBv2 generator, thus, biases in the training data can transfer to our inversion network. For future work, we want to improve the method by transitioning from instant-level to real-time editing capabilities. This enhancement would address current limitations and have a significant impact across various fields.

{
    \small
    \bibliographystyle{ieeenat_fullname}
    \bibliography{main}
}

\maketitlesupplementary

In this supplementary material, we first provide a detailed derivation of the regularization loss used in Stage 2, as outlined in \cref{sec:regu_loss}. Next, we present several additional ablation studies in \cref{sec:ablation}. Finally, we include more quantitative and qualitative results in \cref{sec:quantitative}, and \cref{sec:qualitative}. Then we discuss societal impacts in \cref{sec:soc_impact}.

\section{Derivation of the Regularization Loss in Stage 2}
\label{sec:regu_loss}

We provide a detailed derivation of the gradient of our proposed regularization loss, as defined in Eq.~(8) of the main paper. The regularization loss is formulated as follows:
\begin{equation}
\mathcal{L}_{\text{regu}}^{\text{stage2}} = \mathbb{E}_{t, \hat{\bepsilon}} \left[w(t)\|\bepsilon_\phi(\z_t, t, \c_y) - \hat{\bepsilon}\|^2_2\right] \,,
\label{eq:loss_regu}
\end{equation}
where $\bepsilon_\phi(.)$ is a teacher denoising UNet, here, we use SD 2.1 in our implementation. 

The gradient of the loss w.r.t our inversion network's parameters $\theta$ is computed as:
\begin{equation}
\begin{split}
\nabla_{\theta} \mathcal{L}_{\text{regu}}^{\text{stage2}}\triangleq \mathbb{E}_{t, \hat{\bepsilon}} \left[w(t)(\bepsilon_\phi(\z_t, t, \c_y) - \hat{\bepsilon}\right) 
\\
(\frac{\partial \bepsilon_{\phi}(\z_t, t, \c_y)}{\partial \theta} -\frac{\partial \hat{\bepsilon}}{\partial \theta})]
\label{eq:grad_loss_regu},
\end{split}
\end{equation}
where we absorb all constants into $w(t)$. Expanding the term $\frac{\partial \bepsilon_{\phi}(\z_t, t, \c_y)}{\partial \theta}$, we have:
\begin{equation}
\frac{\partial \bepsilon_{\phi}(\z_t, t, c_y)}{\partial \theta} = \frac{\partial \bepsilon_{\phi}(\z_t, t, c_y)}{\partial \z_t} 
\frac{\partial \z_t}{\partial \z} 
\frac{\partial \z}{\partial \theta}.
\label{eq:teacher_term}
\end{equation}

Since $\z$ (extracted from real images) and $\theta$ are independent,  $\frac{\partial \z}{\partial \theta} = 0$, thus, we can turn \cref{eq:grad_loss_regu} into:
\begin{align}
\nabla_{\theta} \mathcal{L}_{\text{regu}}^{\text{stage2}} & \triangleq \mathbb{E}_{t, \hat{\bepsilon}} \left[w(t)(\bepsilon_\phi(\z_t, t, \c_y) - \hat{\bepsilon}) 
(-\frac{\partial \hat{\bepsilon}}{\partial \theta})\right] \\
& = \mathbb{E}_{t, \hat{\bepsilon}} \left[w(t)(\hat{\bepsilon} - \bepsilon_\phi(\z_t, t, \c_y)) 
\frac{\partial \hat{\bepsilon}}{\partial \theta}\right],
\label{eq:final_regu_grad}
\end{align}
which has the opposite sign of the SDS gradient w.r.t $\z$ loss as discussed in the main paper.

\begin{figure}[t]
    \centering
    \includegraphics[width=0.9\columnwidth]{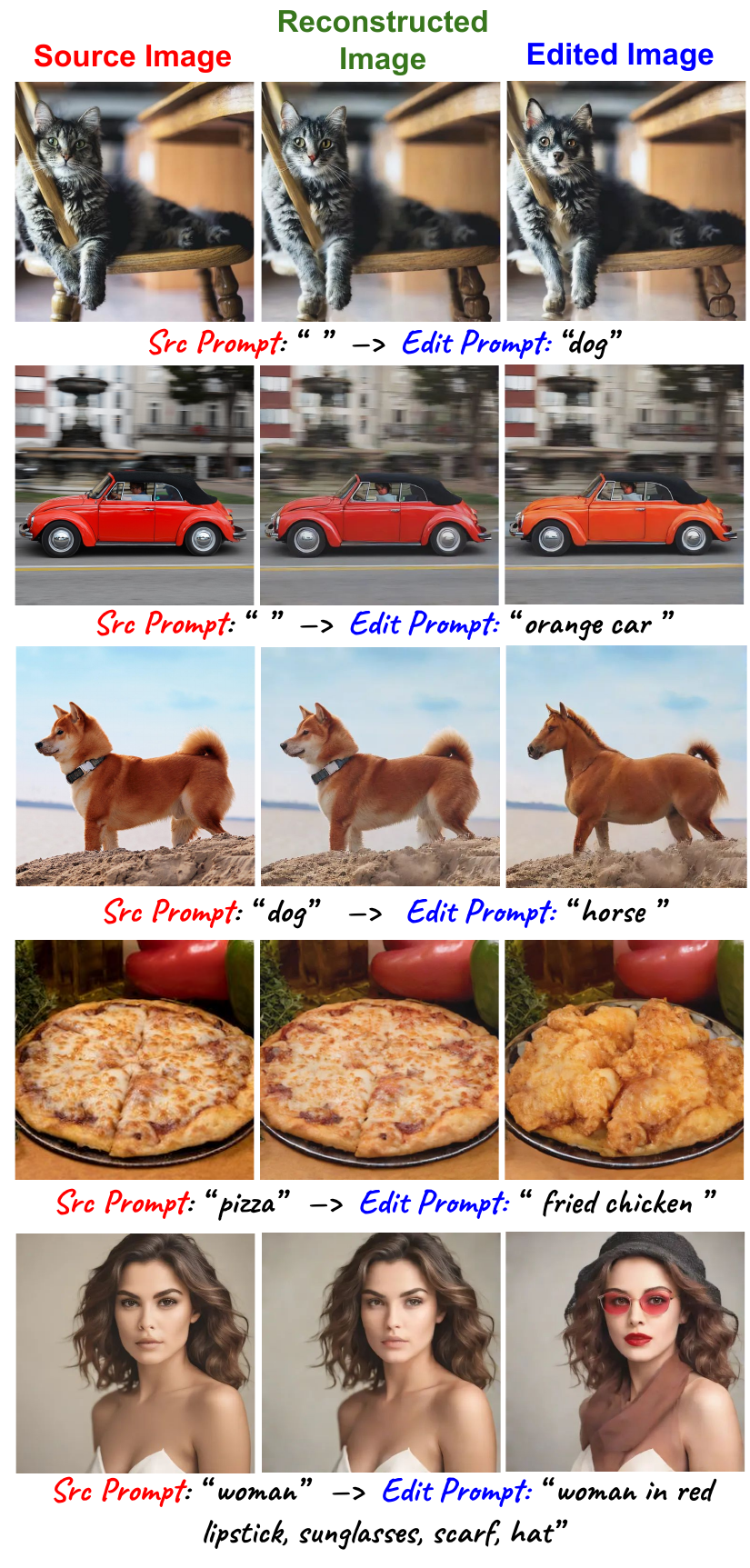}
    \caption{\textbf{Edit images with flexible prompting.} SwiftEdit achieves satisfactory reconstructed and edited results with flexible source and edit prompt input (denoted under each image).}
    \label{fig:flexible_prompt}
\end{figure}

\section{Additional Ablation Studies}
\label{sec:ablation}
\myheading{Compatibility of multi-step inversion with one-step text-to-image model.} To showcase the strength of our one-step inversion framework, we test existing inversion techniques on one-step generators. Specifically, we evaluate multi-step methods like DDIM Inversion (DDIMInv) and direct inversion on SBv2. As shown in the first and second row of \cref{tab:quan_rebuttal}, these methods yield lower performance and slower inference time, while SwiftEdit excels with superior results and high efficiency.

\myheading{Combined with other one-step text-to-image models.} 
As discussed in the main paper, our inversion framework is not limited to SBv2 and can be seamlessly integrated with other one-step text-to-image generators. To demonstrate this, we conducted experiments replacing SBv2 with alternative models, including DMD2 \cite{yin2024improved}, InstaFlow \cite{liu2023instaflow}, and SBv1 \cite{nguyen2024swiftbrush}. For these experiments, the architecture and pretrained weights of each generator $\G$ were used to initialize our inversion network in Stage 1. Specifically, DMD2 was implemented using the SD 1.5 backbone, while InstaFlow uses SD 1.5. All training experiments for both stages were conducted on the same dataset, similar to the experiments presented in Tab.~1 of the main paper.

\begin{figure}
    \centering
    \includegraphics[width=\columnwidth]{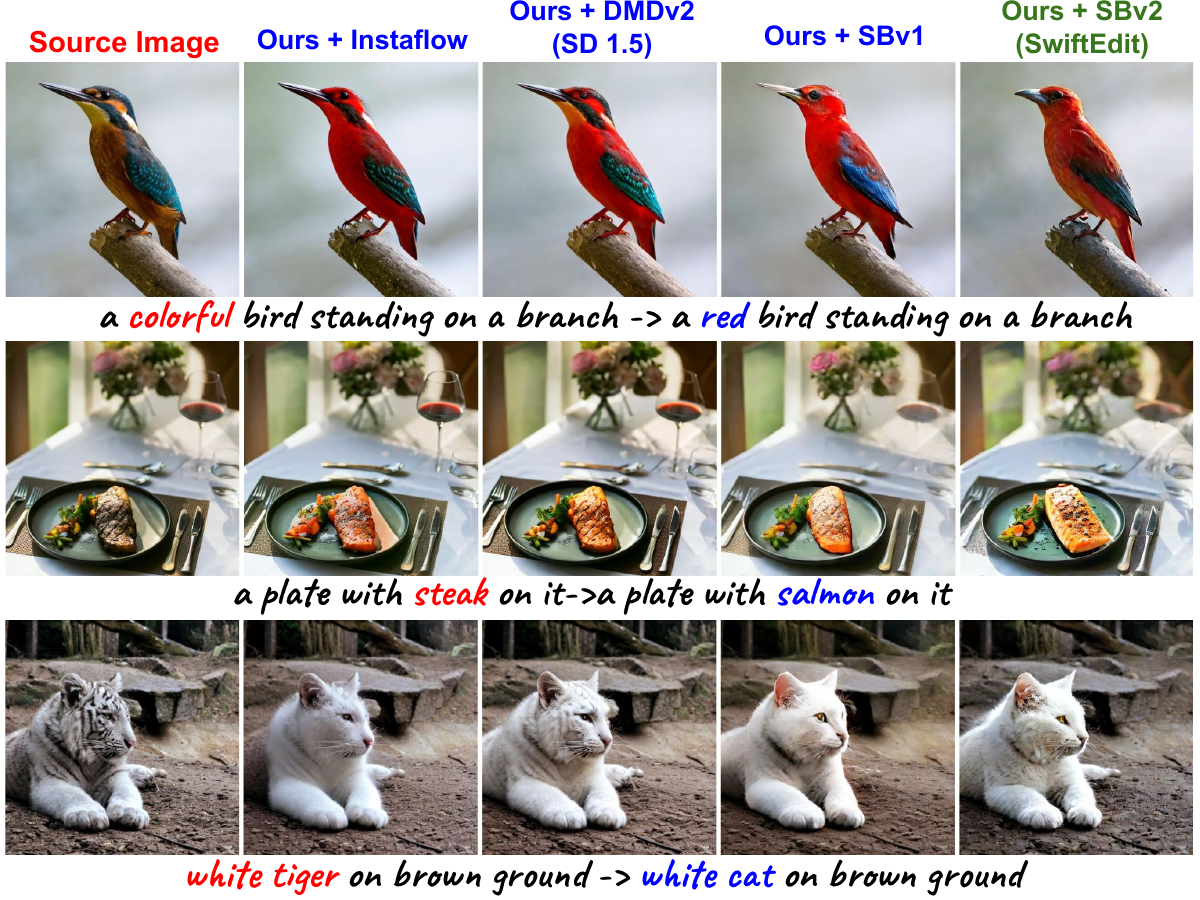}
    \caption{Qualitative results when combining our inversion framework with other one-step text-to-image generation models. }
    \label{fig:edit_with_other_one_step}
\end{figure}

\Cref{fig:edit_with_other_one_step} presents edited results obtained by integrating our inversion framework with different one-step image generators. As shown, these one-step models integrate well with our framework, enabling effective edits. Additionally, quantitative results are provided in \cref{tab:ablation_other_1step}. The results indicate that our inversion framework combined with SBv2 (SwiftEdit) achieves the best editing performance in terms of CLIP-Whole and CLIP-Edited scores, while DMD2 demonstrates superior background preservation.

\begin{table}[t]
\centering
\setlength{\tabcolsep}{2pt}
\adjustbox{width=\columnwidth}{
\begin{tabular}{lcccc}
\toprule
\multirow{1}{*}{\textbf{Model}} & \textbf{PSNR}$\uparrow$
& \textbf{CLIP-Whole}$\uparrow$ & \textbf{CLIP-Edited}$\uparrow$ \\
\midrule
Ours + InstaFlow$^\dagger$ & 24.88 & 24.03 & 20.47\\
Ours + DMD2$^\dagger$ & \textbf{26.08} & 23.35 & 19.84 \\
Ours + SBv1$^\ddagger$  & 25.09 & 23.64 & 19.96\\
\midrule
Ours + SBv2$^\ddagger$ (\textbf{SwiftEdit})&23.33 & \textbf{25.16}& \textbf{21.25}
  
\\
\bottomrule
\end{tabular}
}
\vspace{-5pt}
\caption{Ablation studies on combining our technique with other one-step text-to-image generation models. $\dagger$ means that these models are based on SD 1.5 while $\ddagger$ means that these models are based on SD 2.1.}
\label{tab:ablation_other_1step}
\vspace{-10pt}
\end{table}

\myheading{Two-stage training rationale.}
We provide additional ablation study where we train our network in a single stage using a mixed dataset of synthetic and real images. In particular, we construct a mixed training dataset comprised of: 10,000 synthetic image samples (generated by SBv2 using COCOA prompts), and 10,000 real samples of COCOA dataset. The goal of this experiment is to understand the behavior and advantage of two-stage training compared to single stage training with mixed dataset. As shown in the third row of \cref{tab:quan_rebuttal}, the combined training stage resulted in lower performance across all metrics compared to our two-stage strategy. This highlights the effectiveness of our two-stage strategy.

\begin{table*}[t]
\centering
\setlength{\tabcolsep}{3pt}
\small
\begin{tabular}{lcccccccc} 
\toprule
\multirow{1}{*}{\textbf{Method}} & \textbf{SDis}$\downarrow$ & \textbf{PSNR}$\uparrow$ & \textbf{LPIPS}$\downarrow$ & \textbf{MSE}$\downarrow$ & \textbf{SSIM} $\uparrow$ & \textbf{CLIP-W} $\uparrow$ & \textbf{CLIP-E}$\uparrow$ & \textbf{Time (s)$\downarrow$} \\ 
\midrule
\textbf{DirectInv + SBv2}  & 0.050 & 15.5 & 0.25 & 0.003 & 0.65 & 24.3 & 20.3 & 9.25 \\
\textbf{DDIMInv + SBv2}  & 0.060 & 14.4 & 0.29 & 0.004 & 0.63 & 22.7 & 19.7 & 3.85
\\
\textbf{SwiftEdit (Mixed Training)} & 0.005 & 22.5 & 0.09 & 0.0008 & 0.79 & 23.5 & 19.3 & 0.23 \\
\midrule
\textbf{SwiftEdit (Ours)} & \textbf{0.001} & \textbf{23.3} & \textbf{0.08} & \textbf{0.0006} & \textbf{0.81} & \textbf{25.2} & \textbf{21.3} & \textbf{0.23} \\
\bottomrule
\end{tabular}
\caption{Comparison of SwiftEdit with other settings on PieBench.}
\label{tab:quan_rebuttal}
\end{table*}

\begin{figure}[t]
    \centering
    \begin{subfigure}[b]{\columnwidth}
        \centering
        \includegraphics[width=\columnwidth]{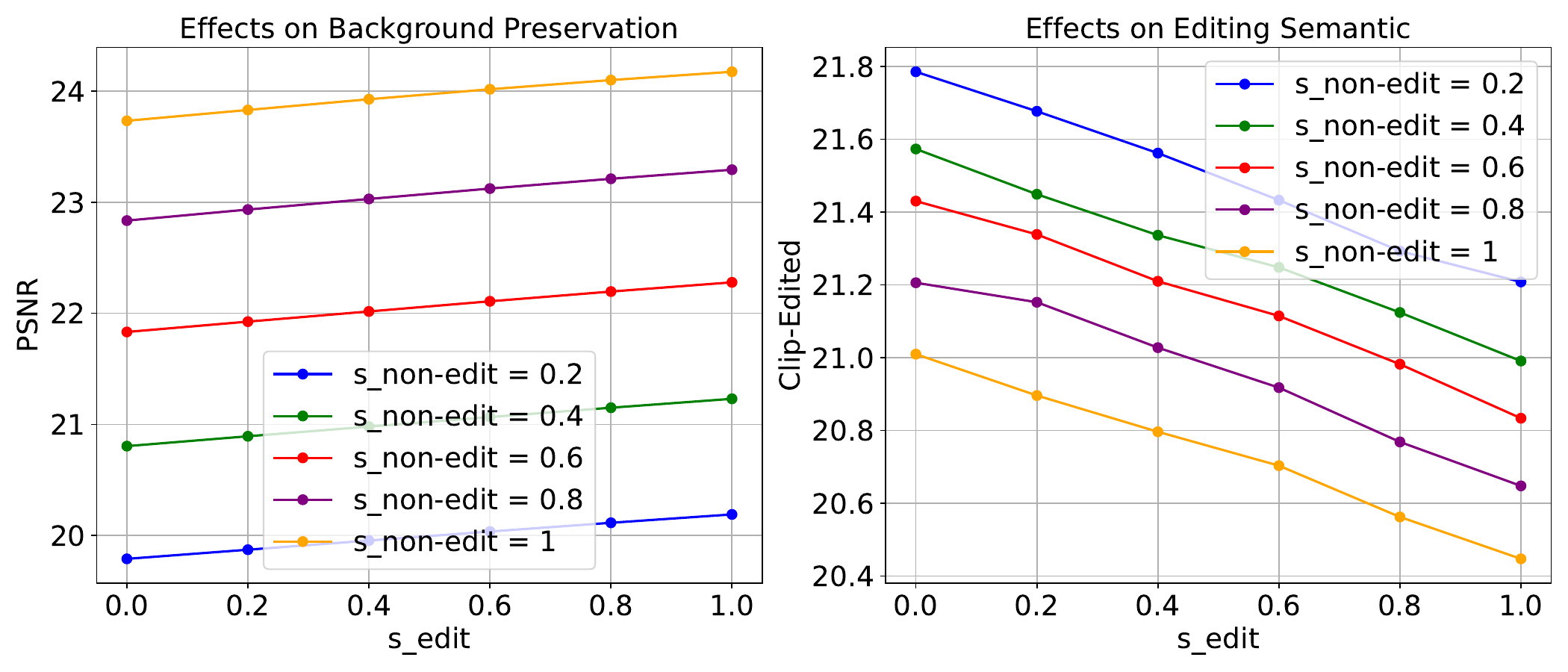}
        \caption{Varying $s_\text{edit}$ scale at different levels of $s_\text{non-edit}$ with default $s_y=2$.}
        \label{fig:vary_sedit}
    \end{subfigure}
    \begin{subfigure}[b]{\columnwidth}
        \centering
        \includegraphics[width=\columnwidth]{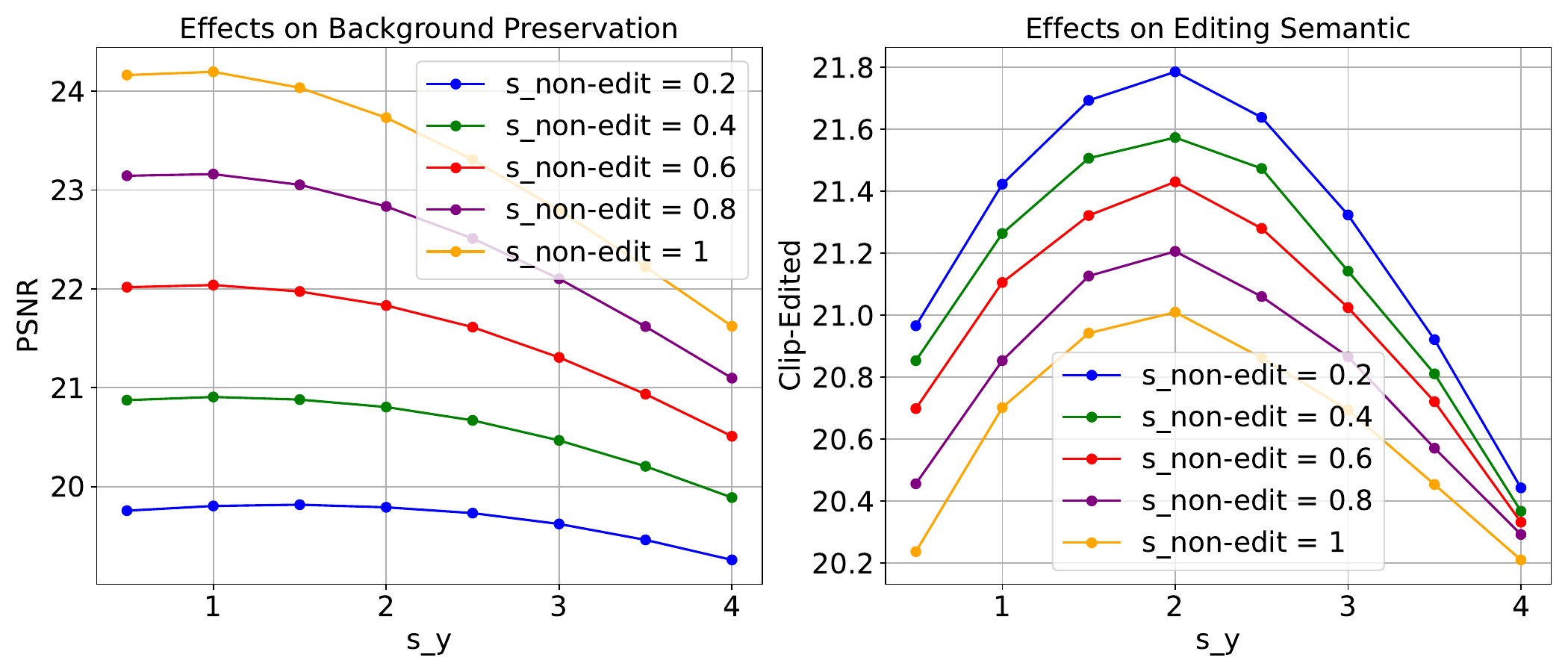}
        \caption{Varying $s_y$ scale at different levels of $s_\text{non-edit}$ with default $s_\text{edit}=0$.}
        \label{fig:vary_sta}
    \end{subfigure}
    \vspace{-5mm}
    \caption{Effects on background preservation and editing semantics while varying $s_\text{edit}$ and $s_y$ at different levels of $s_\text{non-edit}$.}
    \vspace{-4mm}
    \label{fig:vary_two_scales}
\end{figure}

\myheading{Varying scales.}
To better understand the effect of varying scales used in Eq.~(9) in the main paper, we present two comprehensive plots evaluating the performance of SwiftEdit on 100 random test samples from the PieBench benchmark. Particularly, the plots depict results for varying $s_\text{edit} \in \{0, 0.2, 0.4, 0.6, 0.8, 1\}$ (see \cref{fig:vary_sedit}) or $s_\text{y} \in \{0.5, 1, 1.5, 2, 2.5, 3, 3.5, 4\}$ (see \cref{fig:vary_sta}) at different levels of $s_\text{non-edit} \in \{0.2, 0.4, 0.6, 0.8, 1\}$. As shown in \cref{fig:vary_sedit}, it is evident at different levels of $s_\text{non-edit}$ that lower $s_\text{edit}$ generally improves editing semantics (CLIP-Edited scores) but slightly compromises background preservation (PSNR). Conversely, higher $s_\text{y}$ can enhance prompt-image alignment (CLIP-Edited scores, \cref{fig:vary_sta}), but excessive values ($s_\text{y} > 2$) may harm prompt-alignment result. In all of our experiments, we use default choice of scale parameters setting where we set $s_\text{edit}=0$, $s_\text{non-edit}=1$, and $s_\text{y}=2$.


\begin{figure}[t]
    \centering
    \includegraphics[width=\columnwidth]{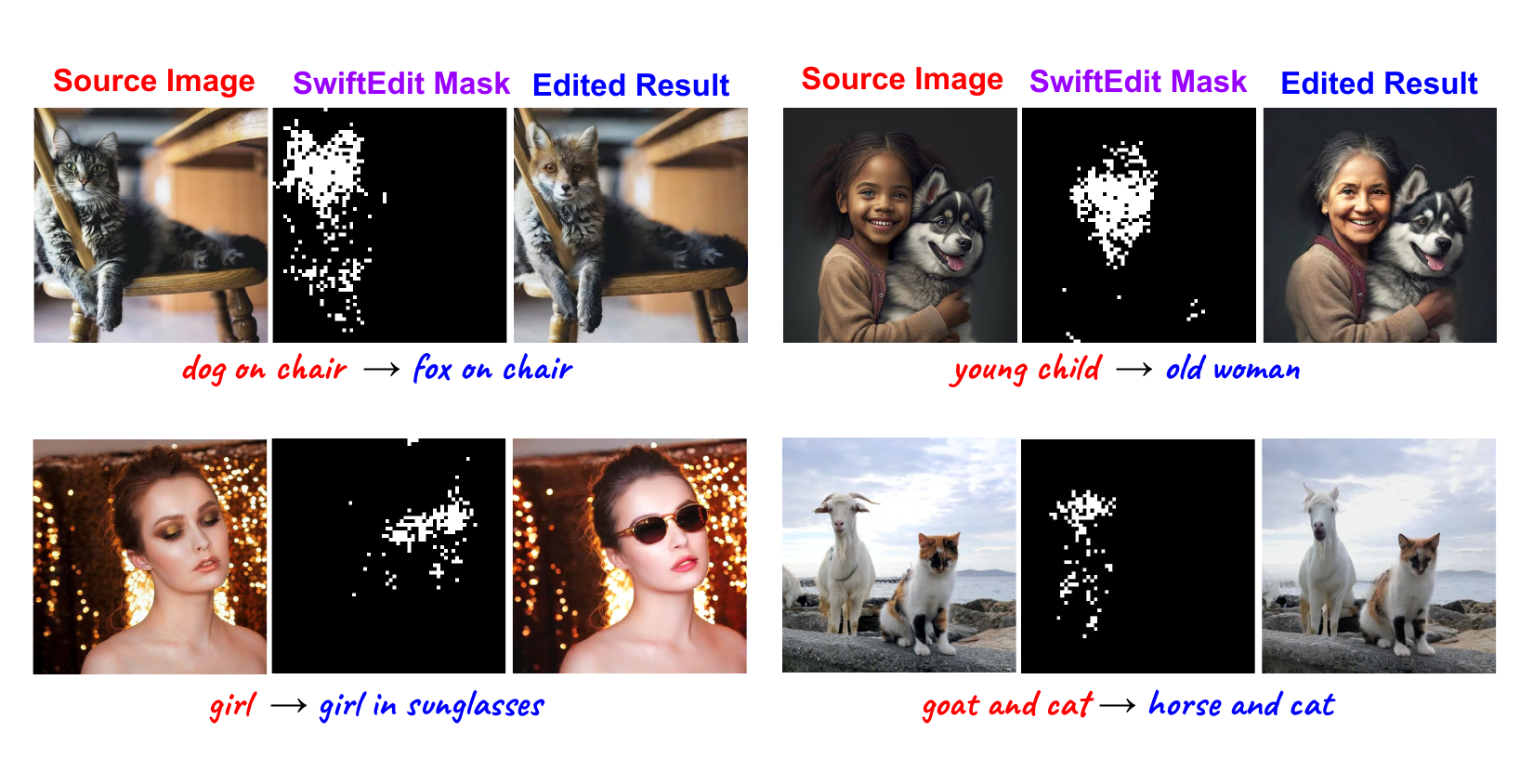}
    \caption{Visualization of our extracted mask along with edited results using guided text described under each image row.}
    \label{fig:additional_mask_vis}
\end{figure}

\begin{figure*}[t]
    \centering
\includegraphics[width=.9\linewidth]{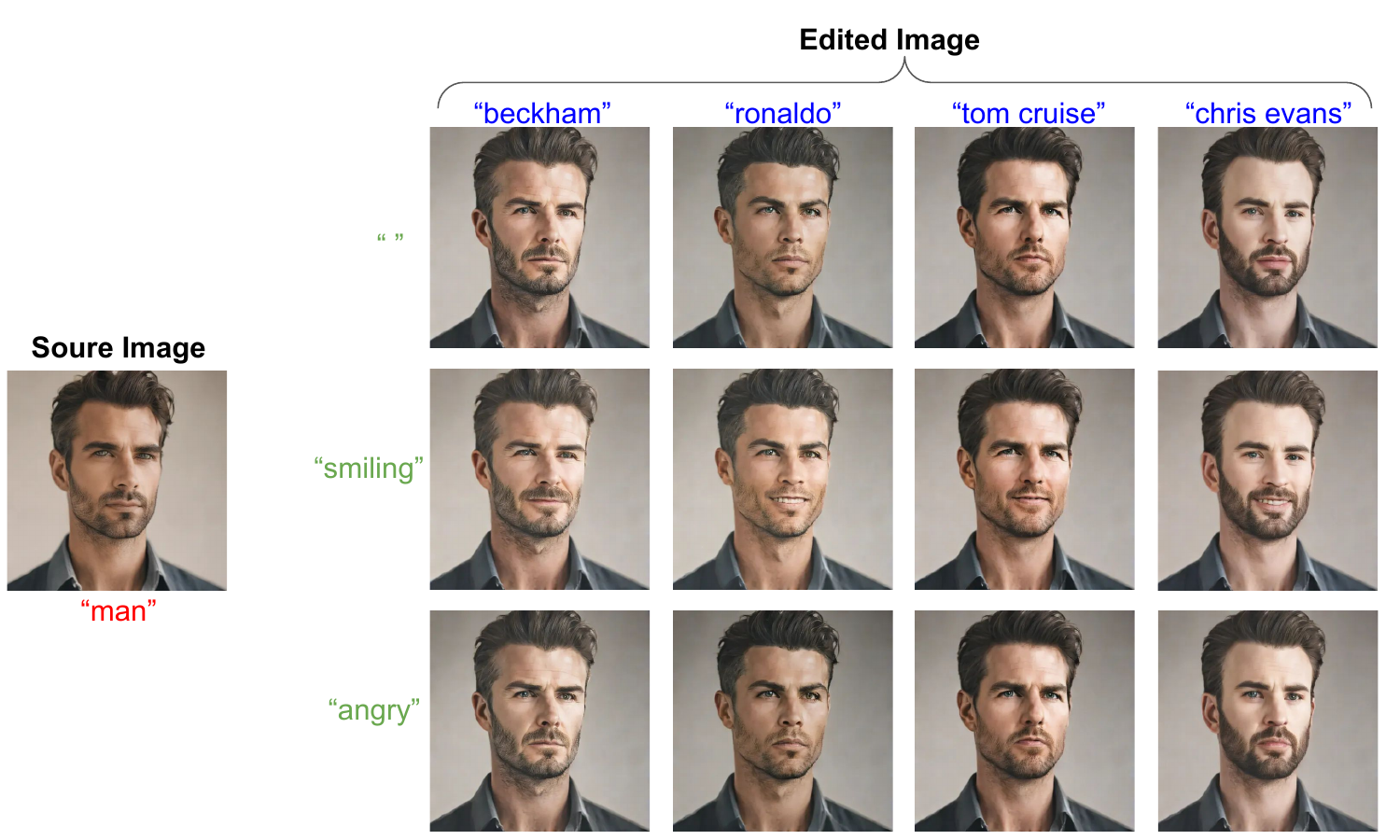}
    \vspace{-2.3mm}
    \caption{\textbf{Face identity and expression editing via simple prompts}. Given a portrait input image, SwiftEdit can perform a variety of facial identities along with expression editing scenarios guided by simple text within just \textbf{0.23} seconds.}
    \label{fig:edit_face}
\end{figure*}

\section{More Quantitative Results}
\label{sec:quantitative}
\begin{table*}[ht]
\centering
\small
\setlength{\tabcolsep}{1pt}
\begin{tabular}{llcccccccc} 
\toprule
\multirow{1}{*}{\textbf{Type}} & \multirow{1}{*}{\textbf{Method}} & \textbf{SDis}$_{\times 10 ^3}$$\downarrow$ & \textbf{PSNR}$\uparrow$ & \textbf{LPIPS}$_{\times 10 ^3}$$\downarrow$ & \textbf{MSE}$_{\times 10 ^4}$$\downarrow$ & \textbf{SSIM}$_{\times 10 ^2}\uparrow$ & \textbf{CLIP-W} $\uparrow$ & \textbf{CLIP-E}$\uparrow$ & \textbf{Time}$\downarrow$\\
\midrule
\multirow{8.9}{*}{\begin{tabular}[c]{@{}l@{}}\textbf{Multi-step}\\\textbf{(50 steps)}\end{tabular}} & DDIM + P2P & 69.43 & 17.87 & 208.80 & 219.88 & 71.14 & \underline{25.01} & 22.44 & 25.98 \\
& NT-Inv + P2P & 13.44 & \textbf{27.03} & \textbf{60.67} & 35.86 & \textbf{84.11} & 24.75 & 21.86 & 134.06 \\ 
\cmidrule{2-10}
& DDIM + MasaCtrl & 28.38 & 22.17 & 106.62 & 86.97 & 79.67 & 23.96 & 21.16 & 23.21 \\
& Direct Inversion + MasaCtrl & 24.70 & 22.64 & 87.94 & 81.09 & 81.33 & 24.38 & 21.35 & 29.68 \\ 
\cmidrule{2-10}
& DDIM + P2P-Zero & 61.68 & 20.44 & 172.22 & 144.12 & 74.67 & 22.80 & 20.54 & 35.57 \\
& Direct Inversion + P2P-Zero & 49.22 & 21.53 & 138.98 & 127.32 & 77.05 & 23.31 & 21.05 & 35.34 \\ 
\cmidrule{2-10}
& DDIM + PnP & 28.22 & 22.28 & 113.46 & 83.64 & 79.05 & 25.41 & \underline{22.55} & 12.62 \\
& Direct Inversion + PnP & 24.29 & 22.46 & 106.06 & 80.45 & 79.68 & 25.41 & \textbf{22.62} & 12.79 \\
\cmidrule{2-10}
& InstructPix2Pix & 57.91 & 20.82 & 158.63 & 227.78 & 76.26 & 23.61 & 21.64 & 3.85 \\
& InstructDiffusion & 75.44 & 20.28 & 155.66 & 349.66 & 75.53 & 23.26 & 21.34 & 7.68 \\
\midrule
\multirow{3}{*}{\begin{tabular}[c]{@{}l@{}}\textbf{Few-steps}\\\textbf{(4 steps)}\end{tabular}} & ReNoise (SDXL Turbo) & 78.44 & 20.28 & 189.77 & 54.08 & 70.90 & 24.30 & 21.07 & 5.10 \\
& TurboEdit & 16.10 & 22.43 & 108.59 & 9.48 & 79.68 & 25.50 & 21.82 & 1.31 \\
& ICD (SD 1.5) & \textbf{10.21} & \underline{26.93} & \underline{63.61} & \textbf{3.33} & \underline{83.95} & 22.42 & 19.07 & 1.38 \\ 
\midrule
\multirow{2}{*}{\begin{tabular}[c]{@{}l@{}}\textbf{One-step}\\\end{tabular}}
& SwiftEdit (Ours) & \underline{13.21} & 23.33 & 91.04 & 6.58 & 81.05 & 21.16 & 21.25 & \textbf{0.23} \\
& SwiftEdit (Ours with GT masks) & 13.25 & 23.31 & 93.88 & \underline{6.19} & 81.36 & \textbf{25.56} & 21.91 & \textbf{0.23} \\
\bottomrule
\end{tabular}
\vspace{-5pt}
\caption{Quantitative comparison of SwiftEdit against other editing methods with metrics employed from PieBench \cite{ju2023direct}.}.
\label{tab:sub_quant}
\vspace{-10pt}
\end{table*}
In \cref{tab:sub_quant}, we provide full scores on PieBench of comparison results in Tab. 1, with additional scores related to background preservation such as Structure Distance (SDis), LPIPS, and SSIM. We additionally compare with other training-based image editing methods such as InstructPix2Pix (InstructP2P), and InstructDiffusion (InstructDiff). Unlike these methods, which require multi-step sampling and paired training data, SwiftEdit trains on source images alone for one-step editing. As shown, SwiftEdit outperforms both in quality and speed, thanks to its efficient one-step inversion and editing framework.

\section{More Qualitative Results}
\label{sec:qualitative}

\myheading{Self-guided Editing Mask.} In \cref{fig:additional_mask_vis}, we show more editing examples along with self-guided editing masks extracted directly from our inversion network. 

\myheading{Flexible Prompting.}
As shown in \cref{fig:flexible_prompt}, SwiftEdit consistently reconstructs images with high fidelity, even with minimal source prompt input. It operates effectively with just a single keyword (last three rows) or no prompt at all (first two rows). Notably, SwiftEdit performs complex edits with ease, as demonstrated in the last row of \cref{fig:flexible_prompt}, by simply combining keywords in the edit prompt. These results highlight its capabilities as a lightning-fast and user-friendly editing tool.

\myheading{Facial Identity and Expression Editing.} In \cref{fig:edit_face}, given a simple source prompt ``man'' and a portrait image, SwiftEdit can achieve face identity and facial expression editing via a simple edit prompt by just combining \textcolor{ForestGreen}{expression word} (denoted on each row) and \textcolor{blue}{identity word} (denoted on each column).

\begin{figure*}[t]
    \centering
    \includegraphics[width=1\textwidth]{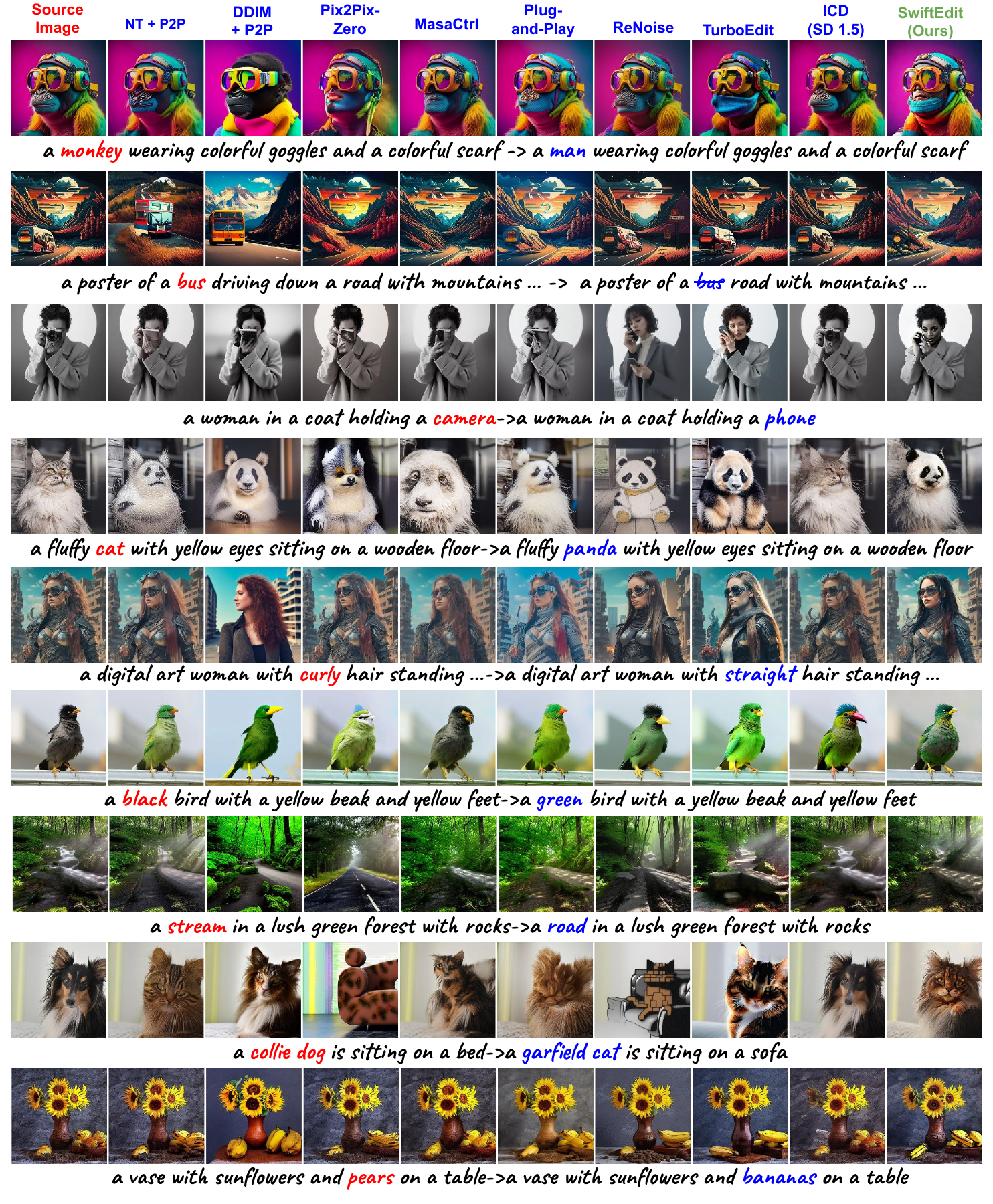}
    \caption{Comparative results on the PieBench benchmark}
    \label{fig:more_qual1}
\end{figure*}

\begin{figure*}[t]
    \centering
    \includegraphics[width=1\textwidth]{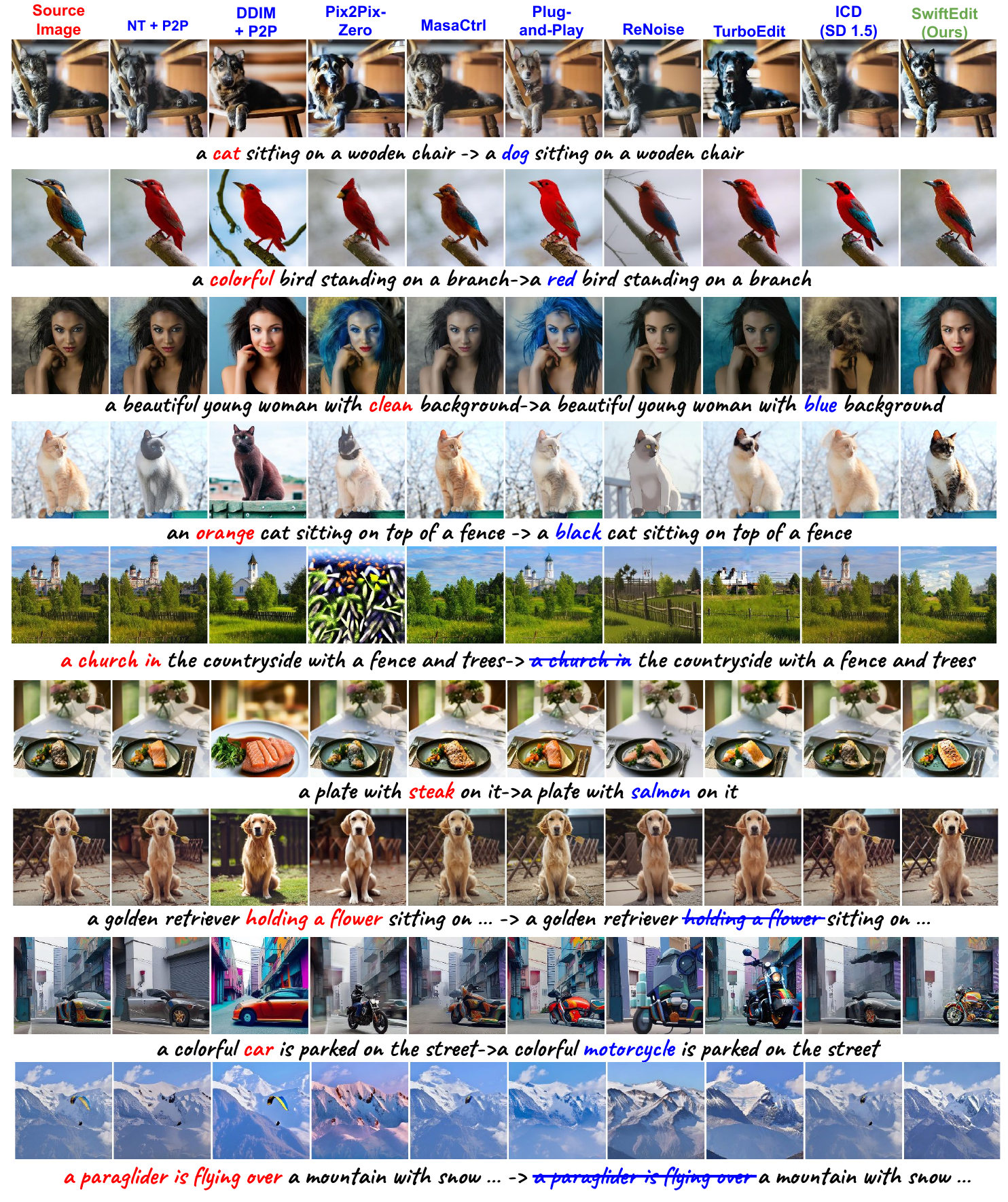}
    \caption{Comparative results on the PieBench benchmark}
    \label{fig:more_qual2}
\end{figure*}

\begin{figure*}[t]
    \centering
    \includegraphics[width=1\textwidth]{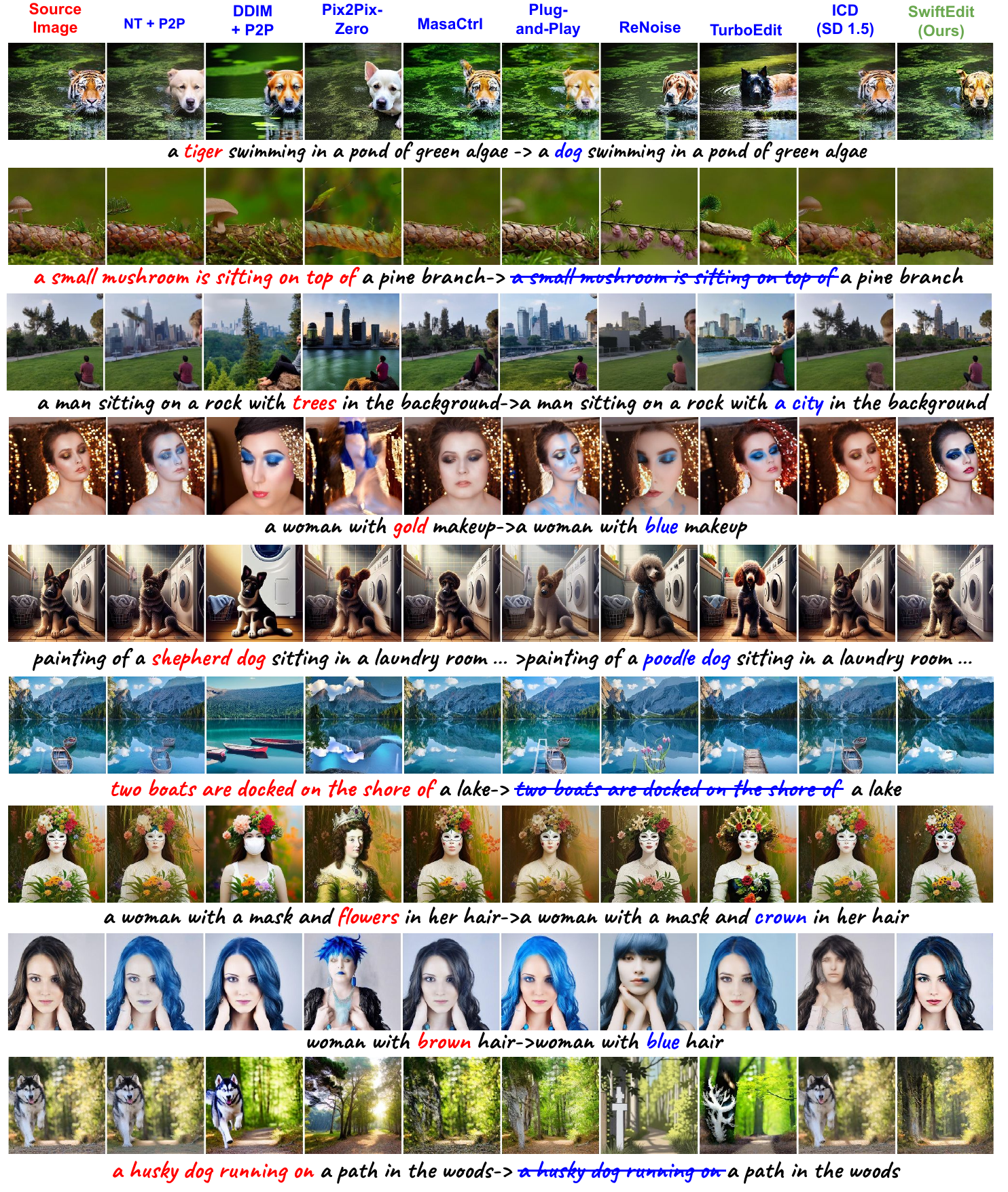}
    \caption{Comparative results on the PieBench benchmark}
    \label{fig:more_qual3}
\end{figure*}

\myheading{Additional Results on PieBench.} In \cref{fig:more_qual1,fig:more_qual2,fig:more_qual3}, we provide extensive editing results compared with other methods on the PieBench benchmark.

\section{Societal Impacts} 
\label{sec:soc_impact}
As an AI-powered visual generation tool, SwiftEdit delivers lightning-fast, high-quality, and customizable editing capabilities through simple prompt inputs, significantly enhancing the efficiency of various visual creation tasks. However, societal challenges may arise as such tools could be exploited for unethical purposes, including generating sensitive or harmful content to spread disinformation. Addressing these concerns are essential and several ongoing works have been conducted to detect and localize AI-manipulated images to mitigate potential misuse.



\end{document}